\documentclass[journal]{IEEEtran}

\usepackage[colorlinks=true,
allcolors=green]{hyperref}
\usepackage{url}
\usepackage{graphics} 
\usepackage{epsfig} 
\usepackage{mathptmx} 
\usepackage{times} 
\usepackage{amsmath} 
\usepackage{amssymb}  
\usepackage[noadjust]{cite}
\usepackage{multirow}
\usepackage{booktabs}
\usepackage{subfigure}
\usepackage{tabularx}
\usepackage{threeparttable}
\usepackage{indentfirst}
\usepackage{algpseudocode}
\usepackage{algorithmicx,algorithm}

\usepackage{makecell}
\usepackage{hyperref}
\usepackage{soul}
\algdef{SE}[DOWHILE]{Do}{doWhile}{\algorithmicdo}[1]{\algorithmicwhile\ #1}

\begin{document}

 \title{Haptics-Enabled Forceps with Multi-Modal Force Sensing: Towards Task-Autonomous Surgery}
	\author{Tangyou Liu, \textit{Student Member, IEEE}, Tinghua Zhang, \textit{Student Member, IEEE}, Jay Katupitiya,\\ Jiaole Wang$^{\ast}$, \textit{Member, IEEE}, and Liao Wu$^{\ast}$, \textit{Member, IEEE}
		\thanks{Research was partly supported by Australian Research
Council under Grant DP210100879, Heart Foundation under Vanguard Grant 106988, and UNSW Engineering GROW Grant PS69063 awarded to Liao Wu, partly by the Science and Technology Innovation Committee of Shenzhen under Grant JCYJ20220818102408018 awarded to Jiaole Wang, and partly by the Tyree IHealthE PhD Top-Up Scholarship awarded to Tangyou Liu. $^{\ast}$Corresponding authors: Jiaole Wang (\textit{wangjiaole@hit.edu.cn}) and Liao Wu (\textit{liao.wu@unsw.edu.au}).}
		\thanks{Tangyou Liu, Jay Katupitiya, and Liao Wu are with the School of Mechanical \& Manufacturing Engineering, The University of New South Wales, Sydney, NSW 2052, Australia.}
		\thanks{Tinghua Zhang and Jiaole Wang are with the School of Mechanical Engineering and Automation, Harbin Institute of Technology, Shenzhen, 518055, China.}
	}

	\maketitle
	
	\begin{abstract}
		Many robotic surgical systems have been developed with micro-sized forceps for tissue manipulation.
		However, these systems often lack force sensing at the tool side and the manipulation forces are roughly estimated and controlled relying on the surgeon's visual perception.
		To address this challenge, we present a vision-based module to enable the micro-sized forceps' multi-modal force sensing. 
		A miniature sensing module adaptive to common micro-sized forceps is proposed, consisting of a flexure, a camera, and a customised target.
		The deformation of the flexure is obtained by the camera estimating the pose variation of the top-mounted target. 
		Then, the external force applied to the sensing module is calculated using the flexure's displacement and stiffness matrix.
		Integrating the sensing module into the forceps, in conjunction with a single-axial force sensor at the proximal end, we equip the forceps with haptic sensing capabilities.
		Mathematical equations are derived to estimate the multi-modal force sensing of the haptics-enabled forceps, including pushing/pulling forces (Mode-I) and grasping forces (Mode-II).
		A series of experiments on phantoms and \textit{ex vivo} tissues are conducted to verify the feasibility of the proposed design and method.
		Results indicate that the haptics-enabled forceps can achieve multi-modal force estimation effectively and potentially realize autonomous robotic tissue grasping procedures with controlled forces.
        A video demonstrating the experiments can be found at \href{https://youtu.be/pi9bqSkwCFQ}{https://youtu.be/pi9bqSkwCFQ}.  
	\end{abstract}

	
	\section{Introduction}
	\label{Section:introduction}
	
\begin{figure*}[t!]
		\centering
		\includegraphics[width=0.98\textwidth]{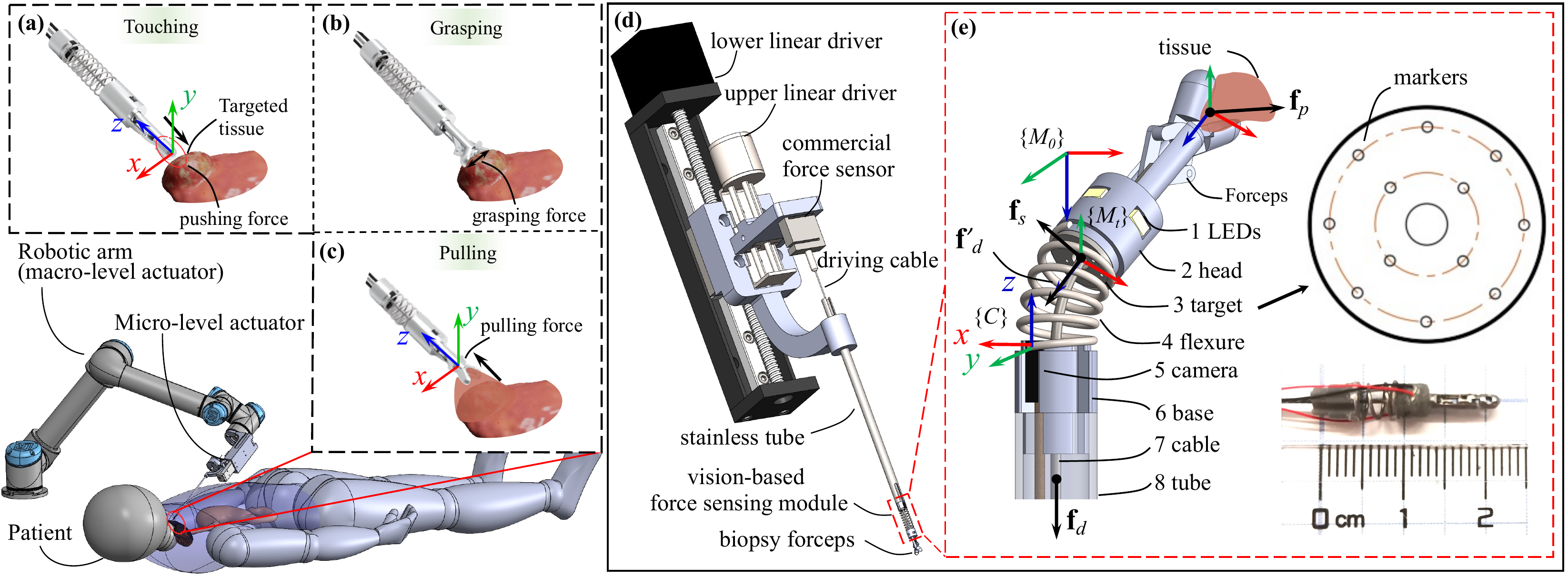}
		\caption{Typical tissue manipulations using a pair of forceps in robotic thyroid tumor treatment, an example scenario of robotic surgery, and the critically required force information.
			(a) A pair of forceps touches the targeted tissue while measuring the pushing forces.
			(b) The forceps grasp the targeted tissue while measuring and controlling the grasping force.
			(c) The forceps pull the grasped tissue while measuring and controlling the pulling force.
			(d) The micro-level actuator used for driving the forceps, where a commercial single-axis force sensor measures the driving force applied to the forceps' driving cable.
			The upper and lower linear drivers are responsible for grasping and pushing/pulling, respectively.
			The lower driver also compensates for the motion introduced by the flexure's deformation when the upper driver grasps tissue.
			(e) The haptics-enabled biopsy forceps, where the forceps' base is concentrically installed to the vison-based sensing module's head (2).
			Two LEDs (1) are mounted in the sensing module's head (2) to provide light source in our prototype, which comes through the target's (3) holes and is captured by the camera (5) mounted on the other end of the flexure (4). 
			The upper-right inset shows the installed target and the holes used as markers.
			The cylindrical base (6) provides a mount to connect to instruments, which is a stainless tube (8) in our prototype.
			Concentric to the module, a channel is reserved for the passage of the forceps' driving cable (7).	
			The lower-right inset shows a prototype with a 4mm diameter and 22mm length. 
			Plot (e) also shows the camera frame \{$C$\}, the target frames of the initial \{$M_0$\} and current \{$M_t$\} states, and the forces applied to the forceps when they push or pull the tissue.
			$\mathbf{f}_d$ is the driving force on the cable and is measured by the proximal commercial force sensor.
            $\mathbf{f}^{\prime}_d$ is the driving force transmitted to the forceps' jaws through the central cable.
            $\mathbf{f}_s$ is the supporting force from the sensing module, and $\mathbf{f}_p$ is the pushing/pulling force from the tissue.}
		\label{Figure:Introduction}	
	\end{figure*}

	\IEEEPARstart{A}{n} increasing number of robotic systems have been developed with miniature instruments to facilitate minimally invasive surgery (MIS) in the past decade \cite{dupont2021decade,dupont2022continuum}.
	For example, micro-sized forceps have been widely adopted in recently developed robotic systems for tissue manipulation in narrow spaces \cite{wu2022camera, feng2022development,cao2022spatial}.
	However, one notable limitation of these systems is their lack of force sensing at the tool side for tissue manipulation, including pushing (Fig.~\ref{Figure:Introduction}(a)), grasping (Fig.~\ref{Figure:Introduction}(b)), and pulling (Fig.~\ref{Figure:Introduction}(c)) forces. 
	
	Force sensing for tissue manipulation is critically needed for two reasons. 
        Firstly, information about these forces can enhance surgeons' operation and decision-making if integrated into the systems appropriately \cite{patel2022haptic}.
	For instance, Talasaz \textit{et al}. \cite{talasaz2016role} verified that a haptics-enabled teleoperation system could perform better during a robot-assisted suturing task.
	Secondly, there is a consensus on the growth of autonomy in surgical robots \cite{yang2017medical} accompanied by more dexterous mechanisms such as snake-like robots \cite{razjigaev2022end,wang2021eccentric}.
	Force sensing at the tool side will play an essential role in the evolution of autonomous robotic surgery \cite{yang2017medical,attanasio2021autonomy}.
	
    Researchers have conducted various investigations to enable force sensing of different grippers for tissue manipulation.
    Zarrin \textit{et al}. \cite{zarrin2018development} fabricated a two-dimensional gripper that can measure grasping and axial forces with two Fiber Bragg grating (FBG) sensors embedded into the gripper's jaws. 
    Using the same principle, Lai \textit{et al}. \cite{lai2021three} developed a gripper that can estimate pulling and lateral forces.
    Although these FBG-based modules usually present a high resolution of force detection, they often suffer from sensitivity to temperature variation, and current compensation methods require complex design \cite{taghipour2019temperature,li2019high, jiang2023fiber}. 
    In addition, highly expensive interrogation systems are needed for these sensors to work. 
    By integrating capacitance into a pair of surgical forceps' two jaws, Kim \textit{et al}. \cite{kim2018sensorized} enabled the forceps with multi-axis force sensing.
    However, capacitance-based methods are often susceptible to temperature and humidity variations, and additional sensing modules or complex design and fabrication are required for compensation \cite{seok2019compensation,sun2015temperature}.
    Moreover, sterilizations are necessary for surgical instruments, such as autoclaving and dry heat procedures, which may harm the sensing capabilities of the tool-side electronic modules \cite{rutala2008guideline}.
    These limitations may invalidate the forceps' force sensing capability when used in practical surgical tasks. 
    In addition, all the above methods require re-machining the forceps' two jaws to install the proposed sensors, making them unsuitable for micro-sized forceps.
	
	Recently, vision-based methods have also been explored for developing low-cost force sensors.
	Ouyang \textit{et al}. \cite{ouyang2020low} developed a multi-axis force sensor with a camera tracking a fiducial marker supported by four compression springs. 
	The displacement of the marker was captured by the camera and converted to force information by multiplying a linear transformation matrix.
	Fernandez \textit{et al.} \cite{fernandez2021visiflex} presented a vision-based force sensor for a humanoid robot's fingers with almost the same principle.
	In addition to providing multi-dimensional force feedback, vision-based force sensing is relatively robust to magnetic, electrical, and temperature variations.
	However, these vision-based sensors developed so far are large in size and can only estimate contact forces.

	Although vision has been adopted for force estimation in surgery, current methods often rely on tissue/organ deformation, movement, and biomechanical properties, and cannot be used as a stand-alone sensing module \cite{haouchine2018vision,fagogenis2019autonomous,chua2022characterization}.
	
	In summary, despite many achievements for gripper force sensing based on various principles in the past few years, force estimation of the micro-sized forceps for MIS, especially their grasping and pulling force measurement, is still challenging. 
	This paper proposes a method to enable multi-modal force sensing of the micro-sized forceps by combining a three-dimensional vision-based force sensing module at the tool side and a single-axis commercial strain gauge sensor at the proximal side, as shown in Fig.~\ref{Figure:Introduction}(d).
    The developed sensing module that can be easily integrated at the tool side is responsible for perceiving the interaction between the forceps and the manipulated tissues (Fig.~\ref{Figure:Introduction}(e)).
    The commercial strain gauge sensor mounted at the proximal side measures the force applied to the forceps' driving cable.
	By integrating these pieces of sensed information, mathematical equations are derived to estimate the forceps' touching, grasping, and pulling forces.

	The main contributions of this article are two-fold:
 \begin{enumerate}
	\item A vision-based force-sensing module that can be easily integrated into micro-sized forceps is developed. 
	An algorithm is designed to calculate the force applied to the sensing module with a registration method for tracking and estimating the pose of the sensing module's target. 
	\item A design of haptics-enabled forceps is further proposed by assembling the widely used biopsy forceps with the developed sensing module embedded at the tool end and a single-axis strain gauge sensor mounted at the proximal end. Mathematical formulae are derived to estimate the multi-modal force sensing of the haptics-enabled forceps, including pushing/pulling force (Mode-I) and grasping force (Mode-II).
\end{enumerate}
 These contributions are validated by various carefully designed experiments on phantoms and \textit{ex vivo} tissues.
    Results indicate that the haptics-enabled forceps can achieve multi-modal force estimation effectively and potentially realize autonomous robotic tissue grasping procedures with controlled forces.

	
	\section{Haptics-Enabled Micro-Sized Forceps}
	\label{Section:sensor_development}

	\subsection{Design of A Vision-Based Force Sensing Module}
	\label{Subsection:sensor_design}
	\subsubsection {Overall Structure}
	The design of the  sensing module (Fig.~\ref{Figure:Introduction}(e)) has followed two requirements: 1) small in size, 2) adaptive to micro-sized forceps, for example, the biopsy forceps.
	The module is designed in a cylindrical shape to minimize anisotropy, and the main components are a flexure, a camera, and a target.
	A central channel is reserved for the passage of instruments.
	The target, supported by the flexure, is manufactured with multiple holes that allow light to pass from a light source, and these holes can then be captured by a camera as markers for the estimation of the target's pose, as depicted in Fig.~\ref{Figure:Introduction}(e).
	The module's size and stiffness can be customized based on the instruments' and applications' needs.
	Here, we assemble a prototype with a diameter of 4mm and a length of 12mm and integrate it in a pair of biopsy forceps, as shown in the lower-right of Fig.~\ref{Figure:Introduction}(e). 
	\subsubsection{Flexure}
	Material properties and geometry play a dominant role in the multidimensional stiffness of the flexure.
	Different approaches have been adopted for flexure design, including the pseudo-rigid-body replacement method \cite{pucheta2010design}, topology optimization \cite{deng2014topology}, etc.
	For force sensing in different surgical tissue manipulations, the flexure should also be changeable according to the task requirements.
	One commercial and readily available flexure is the compression spring \cite{ouyang2020low,fernandez2021visiflex}.
	Although compression springs typically show dominant stiffness in the axial direction, they also exhibit a certain degree of stiffness in lateral directions and can be linearly approximated \cite{keller2011equivalent}.
	Moreover, they can be easily reconfigured by choosing different wire diameters, outer diameters, and coil numbers. 
	For the prototype, we chose a stainless steel compression spring with a wire diameter of 0.5mm, outer diameter of 4mm, rest length of 5mm, and four coils. 
	The selected spring has a force response range of 0$\sim$5N and 0$\sim$2N at axial and lateral displacements of 0$\sim$2mm and 0$\sim$1mm, respectively, which can meet the requirement of most tissue manipulations \cite{golahmadi2021tool}.
	\subsubsection{Camera and Target}
	The camera for the sensing module should have a compact size, sufficient resolution, and surgery compatibility.
	For the prototype, we chose OVM6946 (Omnivision Inc., USA), a medical wafer-level endoscopy camera, which has been adopted in many surgical applications \cite{banach2021visually,cao2022spatial}. 
	It is 1mm in width and 2.27mm in length and has a 400$\times$400 resolution and 30Hz rate.
	To ensure continuous tracking and estimation, we fabricated the target with 12 cycle-distributed 0.15mm-diameter holes used as markers for the camera to track and estimate the target pose, and a central hole is reserved for tools to pass through.
	These holes can also be customized based on the instruments and applications.
	Considering the integration into the biopsy forceps for MIS, the prototype’s target in Fig.~\ref{Figure:Introduction}(e) has a 3.4mm outer diameter and a 0.6mm-diameter central hole for the driving cable.

	\subsection{Force Estimation of Sensing Module}
	\label{Subsection:force_estimation}
	Ideally, for a linear-helix compression spring, the external force $\mathbf{f}_s$ applied to the sensing module's head has an approximated relationship with the spring's displacement \cite{keller2011equivalent} as
	\begin{equation}
		{\mathbf{f}_s} = {{\mathbf{K}}_s}\, {\mathbf{d}_s} = \left[ {\begin{array}{*{20}{c}}
				{{k_x}}&0&0\\
				0&{{k_y}}&0\\
				0&0&{{k_z}}
		\end{array}} \right] 
  \mathbf{d}_s \, ,
		\label{equation:Forcecalculation}
	\end{equation}
	where $\mathbf{f}_s \in \mathbb{R}^{3}$ denotes the external forces, ${\mathbf{K}}_s$ denotes the stiffness matrix of the spring, 
	and $\mathbf{d}_s \in \mathbb{R}^{3}$ is the displacement of the spring's end relative to its relaxed original position.
    In this paper, $\mathbf{d}_s$ is equal to the target's current relative displacement to its initial state.

	\subsubsection{Image-Based Target Pose Estimation}
\begin{figure}[t!]
		\centering
		\includegraphics[width=0.45\textwidth]{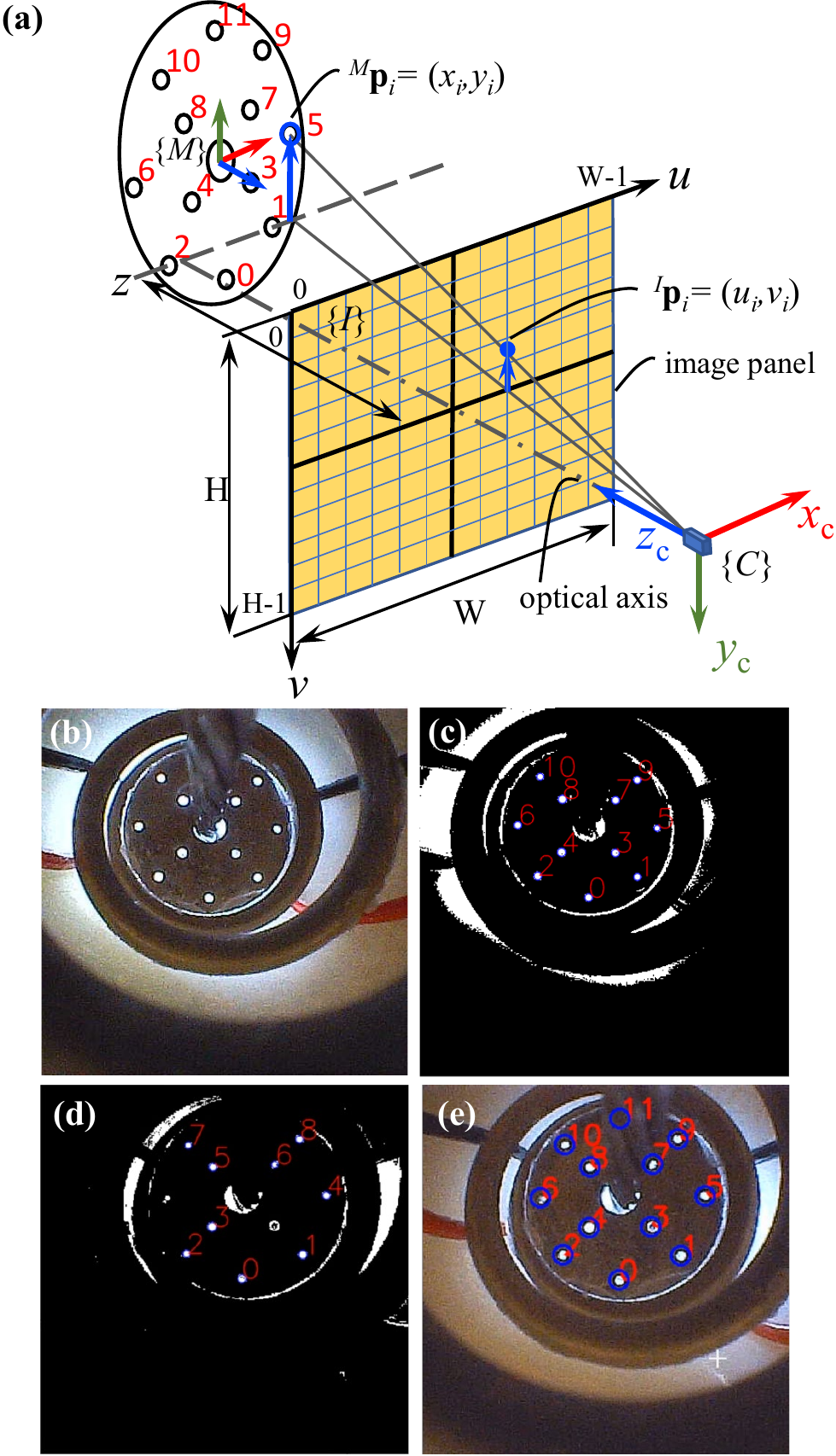}
		\caption{(a) The projection relationship between the $i$-th marker ($^{M}{\mathbf{p}}_{i}$) in the target frame \{\textit{M}\} and its corresponding pixel position ($^{I}{\mathbf{p}}_{i}$) in image frame \{\textit{I}\}. 
			(a) also shows the original index of markers.
			(b) The directly captured image at the initial state. 
			(c) The blob detected result at the initial state based on the filtered image. 
			These detected marker centers are returned for pose estimation.
			(d) The blob detection result at a random pose, where the indexes of detected markers differ, and the points labeled with 3, 6, and 11 in the initial state are lost. 
			(e) The registration result of (e), where markers were registered to their original indexes.}
		\label{Figure:Sensor_design_estimation}
	\end{figure}

	In order to estimate the target pose, each marker needs to be tracked by the camera first.
	However, as the raw image captured by the camera is full of noise, as seen in Fig.~\ref{Figure:Sensor_design_estimation}(b), markers cannot be detected robustly.
	Therefore, a bilateral filter and threshold are adopted to minimize the noise and preserve the boundary in images \cite{singh2019advanced}.
	Then, markers are circled by a parameterized blob detection \cite{kaspers2011blob}, as shown in Fig.~\ref{Figure:Sensor_design_estimation}(c).
	Finally, each marker's central pixel position $^{I}{\mathbf{p}}$ is returned.
	
	The positions of markers $^{M}{\mathbf{P}} \in \mathbb{R}^{2\times 12}$ in the target frame \{\textit{M}\} are known.
	The projection between the $i$-th marker centre $^{M}{\mathbf{p}}_{i} = [x_i, \, y_i]^{\textnormal{T}}$ and its corresponding pixel $^{I}{\mathbf{p}}_{i} = [u_i, \, v_i]^{\textnormal{T}}$ in the image frame \{\textit{I}\}, as shown in Fig.~\ref{Figure:Sensor_design_estimation}(a), can be formulated as 
 	\begin{align}
		^{\mathop{I}}{\tilde{\mathbf{p}}}_{i} 
		& = \frac{1}{z} \, \left[ {\begin{array}{*{20}{c}}
				{\frac{1}{{{\rho _w}}}}&0&{{u_0}}\\
				0&{\frac{1}{{{\rho _h}}}}&{{v_0}}\\
				0&0&1
		\end{array}} \right]\left[ {\begin{array}{*{20}{c}}
				f&0&0&0\\
				0&f&0&0\\
				0&0&1&0
		\end{array}} \right]{}_{M}^{C}{\mathbf{T}}{\,}^{M}{\tilde{\mathbf{p}}_{i}} \notag \\
		& =  \frac{1}{z} \, \underbrace{\left[ {\begin{array}{*{20}{c}}
					{\frac{f}{{{\rho _w}}}}&0&{{u_0}}\\
					0&{\frac{f}{{{\rho _h}}}}&{{v_0}}\\
					0&0&1
			\end{array}} \right]}_{camera \, \, intrinsic \, \, {\mathbf{A}}}
		\left[ {\begin{array}{*{20}{c}}
				r_{11}&r_{12}&t_x\\
				r_{21}&r_{22}&t_y\\
				r_{31}&r_{22}&t_z
		\end{array}} \right]
		\left[ {\begin{array}{*{20}{c}}
				x_i\\
				y_i\\
				1
		\end{array}} \right] \notag \\
		& = \frac{t_z}{z}\, \underbrace{\left[ {\begin{array}{*{20}{c}}
					{h_{11}}&{h_{12}}&{h_{13}}\\
					{h_{21}}&{h_{22}}&{h_{23}}\\
					{h_{31}}&{h_{32}}&1
			\end{array}} \right]}_{homography \, \, \mathbf{H}}  
		\left[ {\begin{array}{*{20}{c}}
				x_i\\
				y_i\\
				1
		\end{array}} \right] \, ,
		\label{equation:projection}
	\end{align} \\ 
	where ${}_{M}^{C}{\mathbf{T}} = \left[ \begin{array}{*{20}{c}}
		\mathbf{r}_1 & \mathbf{r}_2 & \mathbf{r}_3 &\mathbf{t} \\
		0& 0& 0&1
	\end{array}\right] = \left[ {\begin{array}{*{20}{c}}
			r_{11}&r_{12}&r_{13}&t_x\\
			r_{21}&r_{22}&r_{23}&t_y\\
			r_{31}&r_{32}&r_{33}&t_z\\
			0&0&0&1
	\end{array}} \right]$ is the transformation from camera frame \{\textit{C}\} to marker frame \{\textit{M}\},
	$^{M}\tilde{\mathbf{p}}_{i} = [x_i, \, y_i, \, 0, \, 1]^{\textnormal{T}}$ is the homogeneous form of $^{M}{\mathbf{p}}_{i}$,
	$^{I}\tilde{\mathbf{p}}_{i} = [u_i, \, v_i, \, 1]^{\textnormal{T}}$ is the homogeneous form of $^{I}{\mathbf{p}}_{i}$,
	$i\in[0,n]$ ($n=11$ for the prototype), 
	$f$ is the camera's focal length, $\rho_w$ and $\rho_h$ are the width and height of each pixel, respectively, 
 $z = {(h_{31}x_i+h_{32}y_i+1)} \, t_z$, 
 and the camera intrinsic matrix $\mathbf{A}$ is obtained via calibration with MATLAB camera toolbox \cite{fetic2012procedure}.

	According to the transformation relationship shown in Fig.~\ref{Figure:Sensor_design_estimation}(a), the current target pose related to its initial state ${}^{M_0}_{M_t}\mathbf{T}$ can be formulated as
	\begin{equation}
		{}^{M_0}_{M_t}\mathbf{T} = ({}^{C}_{M_0}\mathbf{T})^{-1}{\, \,}^{C}_{M_t}\mathbf{T} \, .
		\label{equation:markerpose}
	\end{equation} 
	The translation of ${}^{M_0}_{M_t}\mathbf{T}$ is equal to the displacement $\mathbf{d}_s$ in (\ref{equation:Forcecalculation}).
	Then, the question becomes to get ${}^{C}_{M_0}\mathbf{T}$ and ${}^{C}_{M_t}\mathbf{T}$.
	The relationship between ${}^{C}_{M}\mathbf{T}$ and homography matrix $\mathbf{H}$ can be formulated as
	\begin{align}
		\left\{\begin{array}{l}
			\mathbf{r}_1 = {t_z}\mathbf{A}^{-1}[h_{11}, h_{21}, h_{31}]^{\rm T}\\
			\mathbf{r}_2 = {t_z}\mathbf{A}^{-1}[h_{12}, h_{22}, h_{32}]^{\rm T}\\
			\mathbf{t} = {t_z}\mathbf{A}^{-1}[h_{13}, h_{23}, h_{33}]^{\rm T}\\
			\mathbf{r}_3 = \mathbf{r}_1 \times \mathbf{r}_2
		\end{array}\right. \, .
		\label{equation:RandT}
	\end{align}
	Because $\mathbf{r}_1$ is a unit vector, $t_z$ can be calculated by $t_z = \frac{1}{||\mathbf{A}^{-1}[h_{11}, h_{21}, h_{31}]^{\rm T}||}$.
	Therefore, to estimate ${}^{C}_{M_0}\mathbf{T}$ and ${}^{C}_{M_t}\mathbf{T}$, we only need to solve the homography matrix $\mathbf{H}^{0}$ and $\mathbf{H}^{t}$ of the initial and current states.

	Because each homography matrix $\mathbf{H}$ has eight variables, at least four independent projections between $^{M}{\mathbf{p}}_{i}$ and $^{I}{\mathbf{p}}_{i}$ are needed to solve it, and each pair's relationship can be formulated as
	\begin{small}
		\begin{equation}
			{\left[ {\begin{array}{*{20}{c}}
						{{}^M{p_{i,x}}}&0\\
						{{}^M{p_{i,y}}}&0\\
						1&0\\
						0&{{}^M{p_{i,x}}}\\
						0&{{}^M{p_{i,y}}}\\
						0&1\\
						{ - {}^M{p_{i,x}}{}^{I}{p_{i,u}}}&{ - {}^M{p_{i,x}}{}^{I}{p_{i,v}}}\\
						{ - {}^M{p_{i,y}}{}^{I}{p_{i,u}}}&{ - {}^M{p_{i,y}}{}^{I}{p_{i,v}}}
				\end{array}} \right]^{\rm T}} \, \left[ {\begin{array}{*{20}{c}}
					{{h_{11}}}\\
					{{h_{12}}}\\
					{{h_{13}}}\\
					{{h_{21}}}\\
					{{h_{22}}}\\
					{{h_{23}}}\\
					{{h_{31}}}\\
					{{h_{32}}}
			\end{array}} \right] = \left[ {\begin{array}{*{20}{c}}
					{{}^{I}{p_{i,u}}}\\
					{{}^{I}{p_{i,v}}}
			\end{array}} \right] \, ,
			\label{equation:homography}
		\end{equation}
	\end{small}
    where $({}^M{p_{i,x}},{}^M{p_{i,y}})$ and $({}^{I}{p_{i,u}},{}^{I}{p_{i,v}})$ are coordinates of the $i$-th marker in the frame \{\textit{M}\} and its projection in the frame \{\textit{I}\}.
	Based on this relationship and four pairs of projections, a set of equations can be established to solve $\mathbf{H}$.
	Substituting the solved $\mathbf{H}$ into (\ref{equation:RandT}), ${}^{C}_{M}\mathbf{T}$ can be further obtained.

	\subsubsection{Marker Robust Registration}
	Ideally, through substituting four detected $^{I}{\mathbf{p}}_{i}$ and their corresponding $^{M}{\mathbf{p}}_{i}$ into (\ref{equation:homography}) we can calculate the homography matrix $\mathbf{H}$ for each state.
	However, under external forces, the pose variation of the target causes the occlusion or halation of some markers and results in wrong correspondence between the image and the marker.
	As a result, the corresponding relationship between $^{I}{\mathbf{p}}_{i}$ and $^{M}{\mathbf{p}}_{i}$ changes over time, which invalidates the direct use of (\ref{equation:homography}).
	Therefore, a registration process is required to label the currently detected markers' pixel positions $^{I}{\mathbf{P}}^{t}$ to their original indexes, as shown in Fig.~\ref{Figure:Sensor_design_estimation}(a), and we define this registered result as $^{I}{\mathbf{P}}^{t,r}$. 
    
	At the initial state $t = 0$, the target is installed perpendicularly to the optical axis.
	For the prototype we built, all the markers can be detected except the 11$^{th}$ marker (top marker) that is occluded by the channel, as shown in Fig.~\ref{Figure:Sensor_design_estimation}(b).
	Therefore, we define $^{M}{\mathbf{P}^0}$ by removing the 11$^{th}$ marker from $^{M}{\mathbf{P}}$ as the marker set at the initial state, and the projection $\mathbf{H}^{0}$ between $^{M}{\mathbf{P}^0}$ and $^{I}{\mathbf{P}}^{0}$ can be obtained according to (\ref{equation:homography}). 
	Substituting $^{M}{\mathbf{p}}_{11}$ and $\mathbf{H}^{0}$ back to (\ref{equation:homography}), we can estimate $^{I}{\mathbf{p}}^{0}_{11}$.
	Then, the complete pixel-position array of the initial state $^{I}{\mathbf{P}}^{0,c}$ can be obtained by appending $^{I}{\mathbf{p}}^{0}_{11}$ to $^{I}{\mathbf{P}}^{0}$.

	For the current state $t$, the pixel positions of the detected markers $^{I}{\mathbf{P}}^{t} \in \mathbb{R}^{2\times m}$ ($m$ is the number of detected markers, $m\leqslant n$) are returned by the blob detection.
	Nevertheless, these detected markers are directly labelled from bottom to top in the image.
    For example, the blob detection result of a random pose is shown in Fig.~\ref{Figure:Sensor_design_estimation}(d). 
	Assuming $^{I}{\mathbf{P}}^{t-1,c}$ is known, we calculate the L1 norm of the distance vector between each $^{I}{\mathbf{p}}^{t}_{i}$ and $^{I}{\mathbf{p}}^{t-1,c}_{j}$, $i \in [1,m]$ and $j \in [1,n]$, and define it as $e$. 
	The corresponding marker of $^{I}{\mathbf{p}}^{t}_{i}$ in $^{I}{\mathbf{P}}^{t-1,c}$ is the one that results in the minimal $e$, and its index is the desired $j^*$.
	Without loss of generality, the registration problem for the detected \textit{i}-th hole is defined as
	\begin{equation}
		j^{*} = \arg \min_j ||^{I}{\mathbf{p}}^{t}_{i} - \, ^{I}{\mathbf{p}}^{t-1,c}_{j}||_1\, .
		\label{equation:registration} 
	\end{equation} 
	Repeating this registration for each detected marker $^{I}{\mathbf{p}}^{t}_{i}$ and updating $^{I}{\mathbf{p}}^{t,r}_{j^{*}} \leftarrow {}^{I}{\mathbf{p}}^{t}_{i}$, we can obtain $^{I}{\mathbf{P}}^{t,r}$ with the original indexes.
    We define the corresponding marker position set of $^{I}{\mathbf{P}}^{t,r}$ as $^{M}{\mathbf{P}}^{t}$.
    Substituting $^{I}{\mathbf{P}}^{t,r}$ and $^{M}{\mathbf{P}}^{t}$ to (\ref{equation:homography}), the homography matrix of the current state $\mathbf{H}^{t}$ can be calculated.
 
	To ensure the image of the current state can be used as the reference for the next registration, i.e., the $t$+1 state, the complete pixel-position set $^{I}{\mathbf{P}}^{t,c}$ of the current state also should be generated. 
    By substituting undetected markers' coordinates in $^{M}{\mathbf{P}}$ and $\mathbf{H}^t$ to (\ref{equation:homography}), we can estimate their corresponding pixel positions $^{I}{\mathbf{P}}^{t,e}$.
    Then, $^{I}{\mathbf{P}}^{t,c}$ can be obtained by merging $^{I}{\mathbf{P}}^{t,e}$ with $^{I}{\mathbf{P}}^{t,r}$.
	For example, Fig.~\ref{Figure:Sensor_design_estimation}(e) shows the registration result and the generated complete image of Fig.~\ref{Figure:Sensor_design_estimation}(d).
	
	\subsubsection{Flexure Deformation Estimation} 
	The transformation between the target and the camera at the initial state ${}^{C}_{M_0}\mathbf{T}$ can be calculated by substituting $^{I}{\mathbf{P}}^{0}$ and $\mathbf{H}^{0}$ into (\ref{equation:RandT}).
	Similarly, the transformation at the current state ${}^{C}_{M_t}\mathbf{T}$ can be obtained by substituting $^{I}{\mathbf{P}}^{t,r}$ and $\mathbf{H}^{t}$ into (\ref{equation:RandT}).
	Then, the transformation ${}^{M_0}_{M_t}\mathbf{T}$ can be calculated by substituting ${}^{C}_{M_0}\mathbf{T}$ and ${}^{C}_{M_t}\mathbf{T}$ into (\ref{equation:markerpose}), which reflects the deformation of the flexure.

	\subsubsection{Force Calculation}
	The displacement of the spring $\mathbf{d}_s$ is the translation vector of ${}^{M_0}_{M_t}\mathbf{T}$.
	Substituting $\mathbf{d}_s$ and the spring's stiffness matrix $\mathbf{K}_s$ into (\ref{equation:Forcecalculation}), we can estimate the external force $\mathbf{f}_s$.
	Ideally, the stiffness matrix $\mathbf{K}_s$ represents the spring's characteristics that connect the camera and the target. 
	However, this stiffness matrix $\mathbf{K}_s$ should also consider the influence of wires and shelter, as they are parallel to the spring.
	In order to get the practical stiffness matrix $\mathbf{K}_s$, a calibration is carried out on the prototype and detailed in section \ref{Section:experiments}. 
    In summary, the above pose and force estimation method can be summarized as algorithm~\ref{Algorithm:Force_Calculation}.
	
		\begin{algorithm}[t]
		\begin{algorithmic}[1]
			\Require
			$^{M}{\mathbf{P}}$, $\mathbf{A}$, $\mathbf{K}_s$ and \textit{frame}
			\Ensure
			${}^{M_t}_{M_0}\mathbf{T}$, $\mathbf{f}_s$
			\State set $t = 0$;
			\State get $^{I}{\mathbf{P}}^{t}$ via blob detection based on \textit{frame};
			\State get $^{M}{\mathbf{P}}^{t}$ by removing $^{M}{\mathbf{p}_{n-1}}$ from $^{M}{\mathbf{P}}$;
			\State calc $\mathbf{H}^{t}$ by substituting $^{I}{\mathbf{P}}^{t}$ and $^{M}{\mathbf{P}}^{t}$ to (\ref{equation:homography});
			\State estimate $^{I}{\mathbf{P}}^{0}_{n-1}$ by substituting $^{M}{\mathbf{p}}_{n-1}$ and $\mathbf{H}^{0}$ to (\ref{equation:homography});
			\State get $^{I}{\mathbf{P}}^{t,c}$ by appending $^{I}{\mathbf{P}}^{t}_{n-1}$ to $^{I}{\mathbf{P}}^{t}$;
			\State calc ${}^{C}_{M_0}\mathbf{T}$ by substituting $\mathbf{H}^{t}$ to (\ref{equation:RandT});
			\While{\textit{frame}}
			\State $t$++;
			\State get $^{I}{\mathbf{P}}^{t}$ via blob detection based on \textit{frame};
			\For{$i = 0; i< col(^{I}{\mathbf{P}}^{t}); i++$}; 
			\State $e'$ = infinity; $j^{*}$ = infinity;
			\For{$j = 0; j< col(^{I}{\mathbf{P}}^{t-1,c}); j++$};
			\State $e = ||^{I}{\mathbf{p}}^{t}_{i} - {\,} ^{I}{\mathbf{p}}^{t-1,c}_{j}||_1$;
			\If{$e<e'$}
			\State $j^{*}$ = j; $e'$ = $e$;
			\Else
			\State $e'$ = $e$;
			\EndIf
			\EndFor 
			\If{$j^{*} \leq col(^{I}{\mathbf{P}}^{t-1,c})$}	
			\State $^{I}{\mathbf{p}}^{t,r}_{j^{*}} \leftarrow{\,} ^{I}{\mathbf{p}}^{t}_{i}$;	
			\EndIf	
			\EndFor
			\State get corresponding $^{M}{\mathbf{P}^t}$ of $^{I}{\mathbf{P}}^{t,r}$ from $^{M}{\mathbf{P}}$;
			\State calc $\mathbf{H}^{t}$ by substituting $^{I}{\mathbf{P}}^{t,r}$ and $^{M}{\mathbf{P}^t}$ to (\ref{equation:homography});            
			\State get $^{I}{\mathbf{P}}^{t,e}$ by substituting $^{M}{\mathbf{P}}$ and $\mathbf{H}^{t}$ to (\ref{equation:projection});
			\State generate $^{I}{\mathbf{P}}^{t,c}$ by merging $^{I}{\mathbf{P}}^{t,r}$ with $^{I}{\mathbf{P}}^{t,e}$;
			\State calc ${}^{C}_{M_t}\mathbf{T}$ by substituting $\mathbf{H}^{t}$ to (\ref{equation:RandT});
			\State calc ${}^{M_0}_{M_t}\mathbf{T}$ by substituting ${}^{C}_{M_0}\mathbf{T}$ and ${}^{C}_{M_t}\mathbf{T}$ to (\ref{equation:markerpose});
			\State calc $\mathbf{f}_s$ by substituting ${}^{M_0}_{M_t}\mathbf{T}$ and $\mathbf{K}_s$ to (\ref{equation:Forcecalculation});
			\EndWhile
		\end{algorithmic}	
		\caption{Pose and Force estimation of sensing module}
		\label{Algorithm:Force_Calculation}
	\end{algorithm}

	\subsection{Haptics-Enabled Forceps and Force Estimation}
	
	\subsubsection{Micro-level Forceps Actuator}
	\label{Subsection:forceps_design}
	To drive the forceps and conduct experimental verification, we designed a micro-level actuator, as shown in Fig.~\ref{Figure:Introduction}(d), which consists of upper and lower linear drivers that are responsible for grasping and pushing/pulling, respectively.
	The lower driver also compensates for the motion introduced by the flexure's deformation when the upper driver actuates the forceps to grasp tissue.
	The compensation is achieved by moving the lower driver's carrier for the same amount of the flexure's axial deformation estimated by the camera in the reverse direction. 
	A commercial single-axis force sensor (ZNLBS-5kg, CHINO, China), with a resolution of 0.015N, is installed collinearly to the forceps and connected to the driving cable to measure the driving force.

	\subsubsection{Pushing/Pulling Force Estimation (Mode-I)}
	\label{Subsection:forceps_pulling}
	Integrating the sensing module into the forceps, we can enable its haptic sensing, and the haptics-enabled forceps are shown in Fig.~\ref{Figure:Introduction}(e).
	The base of the forceps is installed concentrically to the sensing module's head, and the driving cable passes through the reserved central channel.
	The cylindrical base connects the forceps to its carrying instrument (stainless tube).
	A prototype is also shown in the inset of Fig.~\ref{Figure:Introduction}(e), which has a 4mm diameter and 22mm total length and is used in the following description and experiments.
	The proposed haptics-enabled forceps can be easily integrated into a robotic surgical system with the micro-level actuator, such as the robot presented in Fig.~\ref{Figure:Introduction}(a).
	The forces applied to the forceps when they push or pull a tissue are also shown in Fig.~\ref{Figure:Introduction}(e).
	$\mathbf{f}_p$ is the pushing/pulling force from the interactive tissue.
	$\mathbf{f}_s$ is the elastic force from the spring, which can be directly estimated by (\ref{equation:Forcecalculation}).
	$\mathbf{f}^\prime_d$ is the driving force of the cable at the tool end.
	Ignoring friction, we assume $|\mathbf{f}^\prime_d|$ equals $|\mathbf{f}_d|$ measured by the single-axis force sensor connected to the driving cable at the proximal side.
	Then, $\mathbf{f}^\prime_d$ can be formulated as $\mathbf{f}^\prime_d = {}^{M_0}_{M_t}{}\mathbf{T}\,{}\mathbf{f}_d$. 
	Finally, we can obtain the pushing/pulling force $\mathbf{f}_p$ as
	\begin{equation}
		\mathbf{f}_p = -{}^{M_0}_{M_t}{}\mathbf{T}\,{}\mathbf{f}_d \,-\mathbf{f}_s \, .
		\label{equation:fp} 
	\end{equation} 
    For tissue touching, the contact force can also be estimated by (\ref{equation:fp}) no matter if the forceps are closed or open.

	\subsubsection{Grasping Force Estimation (Mode-II)}
	\label{Subsection:forceps_grasping}
\begin{figure}[t]
		\centering
		\includegraphics[width=0.48\textwidth]{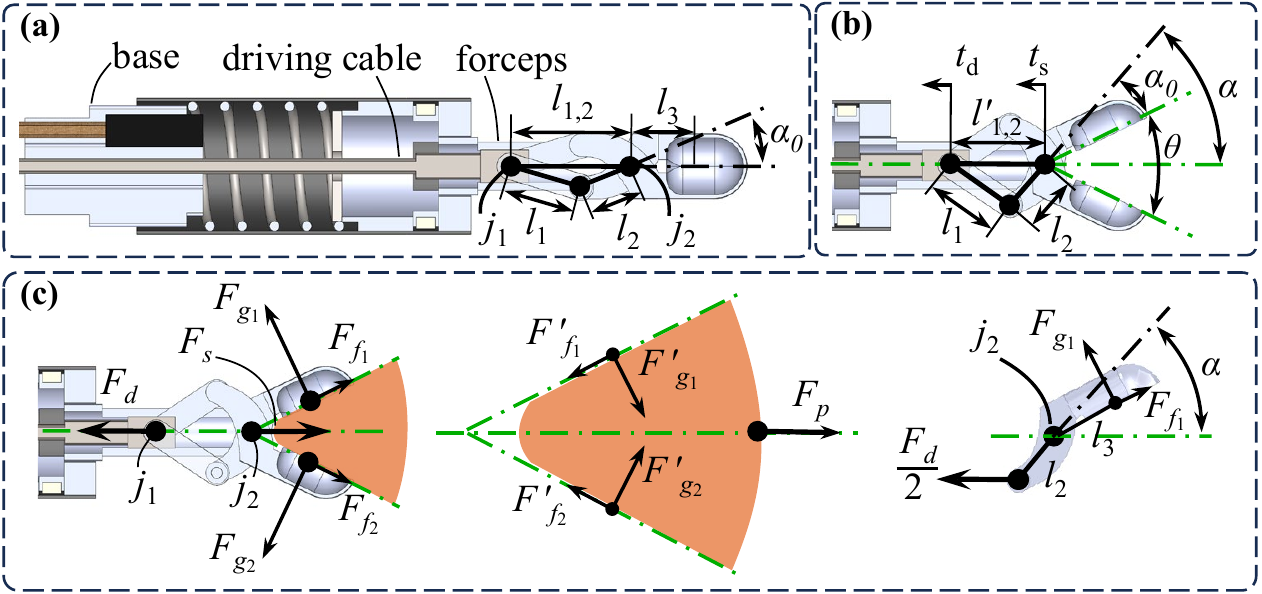}
		\caption{ (a) The sectional view of the haptics-enabled forceps when the two jaws are closed.
			(b) shows a state when the two jaws are open at $\theta$.
			$t_d$ is the movement of the driving cable, while $t_s$ is the transformation of the sensing module's head estimated by the camera. 
			(c) The forces applied to the forceps when they grasp and pull a tissue include driving force $F_d$ from the cable, supporting force $F_s$ from the sensing module, contact force $F_g$ (also called grasping force), and friction force $F_f$ from the tissue.
            The middle inset sketched the forces applied to the tissue, where $F_p$ is the pulling force from the tissue body, $F^\prime_g = F_g$ and $F^\prime_f = F_f$ are from the forceps.
			The right inset shows the forces that generate the momentum of one jaw about joint $j_2$.
			$F_d$ is measured by the proximal force sensor, while the driving distance $t_d$ is measured by the upper driver's encoder of the micro-level actuator.
		}
		\label{Figure:forcpes}	
	\end{figure}

	The estimation of the grasping force is more challenging because it relates to the geometry and driving force applied to the forceps' cable, as depicted in Fig.~\ref{Figure:forcpes}.
	Although the sensing module can estimate the 3D elastic force $\mathbf{f}_s$ and the orientation of $\mathbf{f}^\prime_d$, we mainly consider the situation where the forceps grasp and pull tissue straightly as grasping is performed in the release condition.
	
	The forces applied to the forceps when they grasp and pull a tissue are shown in Fig.~\ref{Figure:forcpes}(c).
    For convenience, we adopt $F_{-}$ to denote the magnitude of force $\mathbf{f}_{-}$ in the following sections.
	$F_s$ is the elastic force from the spring and is directly estimated by (\ref{equation:Forcecalculation}).
	$F_d$ is the driving force applied to the forceps' hinge and measured by the proximal single-axis force sensor.
    $F_g$ and $F_f$ are contact and friction forces from the grasped tissue, as shown in Fig.~\ref{Figure:forcpes}(c).
    For ease of understanding, we define $F_g$ as the gripping force in the following description.

	Assuming the momentum of one jaw about joint $j_2$ is zero, as sketched in the right inset of Fig.~\ref{Figure:forcpes}(c), we can further obtain the relationship between $F_{g_1}$ and $F_d$ as 
	\begin{align}
		F_{g_1} = \frac{F_d \, l_2 \sin{\alpha}}{2\,l_3}\, ,
		\label{equation:F_N}
	\end{align}
	where $\alpha = \frac{\theta}{2} + \alpha_0$, the definitions of $\theta$, $\alpha$, $\alpha_0$, $l_2$, and $l_3$ are shown in Fig.~\ref{Figure:forcpes}(b).
	$\theta$ and $\alpha_0$ can be calculated based on the geometric relationship of the forceps' links.
	Fig.~\ref{Figure:forcpes}(a) shows the forceps' geometry at the initial state, i.e., the forceps are in the closed form, and the flexure is relaxed.
	According to the relationship of forceps' links at the initial state, we can get $\alpha_0$ and formulate it as
	\begin{equation}
		\alpha_0  = \arccos \frac{{l_{1,2}^2 + l_2^2 - l_{1}^2}}{{2\,{l_1}{\,}{l_{1,2}}}} \, ,
	\end{equation}
	where $l_{1,2}$ is the distance between forceps' two joints $j_1$ and $j_2$.
	When the forceps are open at angle $\theta$, the geometry changes to Fig.~\ref{Figure:forcpes}(b).
	At this stage, the angle ${\alpha}$ can be reformulated as
	\begin{equation}
		{\alpha} = \arccos \frac{{l_{1,2}^{\prime2} + l_2^2 - l_1^2}}{{2\,{l_1}\, {l_{1,2}^\prime}}} \, ,
		\label{equation:alpha}
	\end{equation}
	where $ l{_{1,2}^\prime} = l{_{1,2}} + {t}_d - {t}_s$,
	$t_d$ is the movement of the driving cable measured by the upper driver's encoder of the micro-level actuator, and $t_s$ is the transformation of the sensor's head estimated by the sensing module. 
	Moreover, the relationship of $\theta$, $\alpha$, and $\alpha_0$ is
	\begin{equation}
		\theta = 2(\alpha-\alpha_0) \,.
        \label{euquation:theta}
	\end{equation}
    For a pair of forceps, we assume $F_{g_1}$ equals $F_{g2}$, and $F_{f_1}$ equals $F_{f_2}$.
    Then, referring to (\ref{equation:fp}), we can formulate $F_g$ as 
    \begin{equation}
		F_g =  \frac{F_d \, l_2 \sin{\alpha}}{2\,l_3} = \frac{(F_p+F_s) \, l_2 \sin{\alpha}}{2\,l_3} \, .
        \label{euquation:Fg}
  \end{equation}
 

	\section{Experiments \& Discussion}
	\label{Section:experiments}
    In this section, we first present the experiments on evaluating the method for target tracking and calibrating the stiffness matrix for external force estimation.
	Then, the evaluation of two modes of force estimation is described.
	Finally, groups of automatic robotic grasping procedures with the proposed system are illustrated as potential applications.

\begin{figure}[t]
	\centering
	\includegraphics[width=0.48\textwidth]{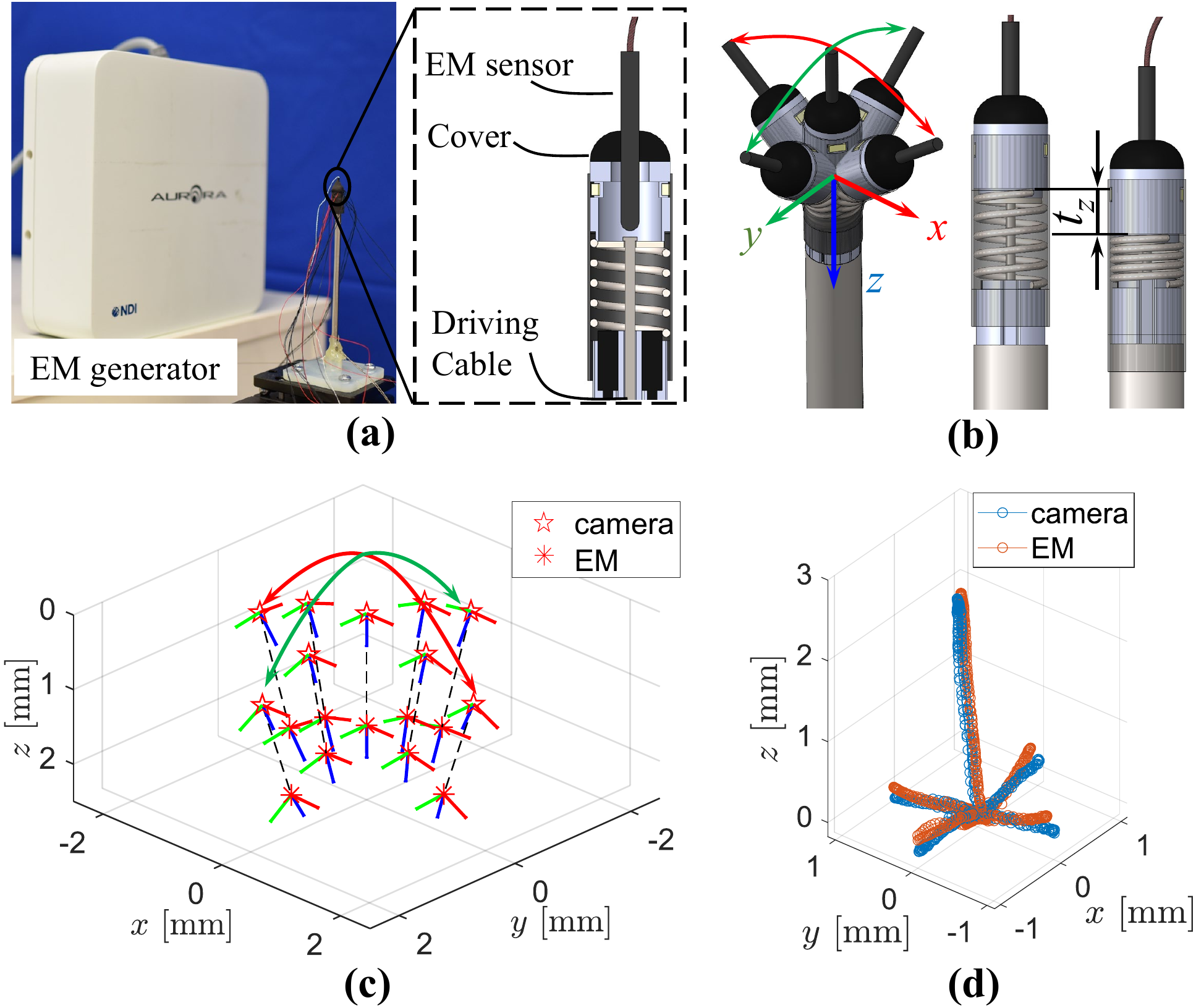}
	\caption{(a) The experimental setup for evaluating pose estimation.
		The inset shows the configuration, where the EM tracking sensor was installed concentrically to the target.
		(b) The sensor's head was moved towards $x$, $y$ and $z$ directions during the experiments, and $t_z$ denotes the transformation along $z$ axis. 
		(c) The orientation comparison between nine pairs of points that were estimated by the EM tracking system ($\star$) and camera ($\ast$). 
		(d) The position comparison between the continued EM tracking and camera estimation results.
	}
	\label{Figure:EM_tracking}	
\end{figure}

	\subsection{Evaluation of Target's Pose Estimation}
	\label{subsection:pose_position_evaluation}
	\subsubsection{Experimental Setup}
	Fig.~\ref{Figure:EM_tracking}(a) shows the experimental setup for pose estimation evaluation with an electromagnetic (EM) tracking system (Northern Digital Inc, Canada) with resolutions less than 0.1mm in position and 0.1$^\circ$ in orientation.
	The EM tracking sensor was installed concentrically to the proposed force sensing module. 
	During the experiment, the sensor's head was moved along $x$ (red arrow), $y$ (green arrow), and $z$ (blue arrow) as indicated in Fig.~\ref{Figure:EM_tracking}(b).

	\subsubsection{Pose Estimation Results}
	The orientation comparison is based on nine pairs of samples distributed in the sensing module's workspace, as plotted in Fig.~\ref{Figure:EM_tracking}(c).
	Here, we use ($\alpha_{C,i}$, $\beta_{C,i}$, $\gamma_{C,i}$) and ($\alpha_{E,i}$, $\beta_{E,i}$, $\gamma_{E,i}$) to denote the pitch, roll, and yaw angles of camera estimation and EM tracking results of the $i$-th pair.
	The max mean and absolute maximum deviations between EM tracking and our estimation result are calculated by $\max \frac{1}{n}({\sum\limits_{i = 1}^n {{\alpha_{C,i}} - {\alpha_{E,i}}}} \, ,{\sum\limits_{i = 1}^n {{\beta_{C,i}} - {\beta_{E,i}}}} \, , {\sum\limits_{i = 1}^n {{\gamma_{C,i}} - {\gamma _{E,i}}} })$ and $\max( \max\limits_{i = 1}^n({\alpha_{C,i}} - {\alpha_{E,i}})\, ,\max\limits_{i = 1}^n({\beta_{C,i}} - {\beta_{E,i}}) \, ,\max\limits_{i = 1}^n({\gamma_{C,i}} - {\gamma_{E,i}}))$, where $n=9$.
	For position evaluation, the EM sensor and target were tracked continuously by the EM tracking system and camera, respectively, and the comparison is shown in Fig.~\ref{Figure:EM_tracking}(d). 
	To calculate the mean and max deviations between the EM and camera-tracking traces, we further calculated the mean ($d_a$) and Hausdorff ($d_h$) distance by
	\begin{align}
		\left\{\begin{array}{l}
			d_a = \max(\frac{1}{n}\sum\limits_{i = 1}^n {{d_{{C_i},E}}} \, ,\frac{1}{m}\sum\limits_{j = 1}^m {{d_{{E_j},C}}})\\
			{d_h} = \max (\mathop {\max }\limits_{i = 1}^n ({d_{C_i,E}}),\mathop {\max }\limits_{j = 1}^m ({d_{E_j,C}}))
		\end{array}
		\right. \, ,
		\label{equation:hausdorff_distance}	
	\end{align}
	where ${d_{{C_i},E}} = \mathop {\min }\limits_{j = 1}^m (||{\mathbf{p}_{C,i} - \mathbf{p}_{E,j}}||_2)$, 
 $\mathbf{p}_{C,i}$ and $\mathbf{p}_{E,j}$ denote positions of the camera estimation of point $i$ and EM tracking of point $j$, 
 ${d_{{E_j},C}} = \mathop {\min }\limits_{i = 1}^n (||{\mathbf{p}_{E,j} - \mathbf{p}_{C,i}}||_2)$,
    $n$=1336 and $m$=2181 are the numbers of points recorded by the camera and EM tracking system, respectively.
As listed in Table \ref{Table:EM_tracking}, the max mean, absolute maximum, and root mean square (RMS) deviations of the orientation are 0.056rad, 0.147rad, and 0.041rad, and those of the position are 0.0368mm, 0.2265mm, and 0.0551mm.
	This indicates that the proposed method can track and estimate the target's pose reliably.

 \begin{table}
	\centering
	\caption{Evaluation of orientation and position deviation}	
	\begin{tabular}{cccc}
		\toprule
		Experiment & Max Mean & Maximum & RMS\\ \toprule
		Orientation & 0.056rad & 0.147rad & 0.041rad\\
		Position & 0.0368mm & 0.2265mm &0.0551mm\\
		\hline
	\end{tabular}
	\label{Table:EM_tracking}
\end{table}

	\subsection{Stiffness Matrix Calibration and Verification}
	\label{Experiments:stiffness_matrix}
	
	\subsubsection{Experimental Setup}
\begin{figure}[t!]
		\centering

  \includegraphics[width=0.45\textwidth]{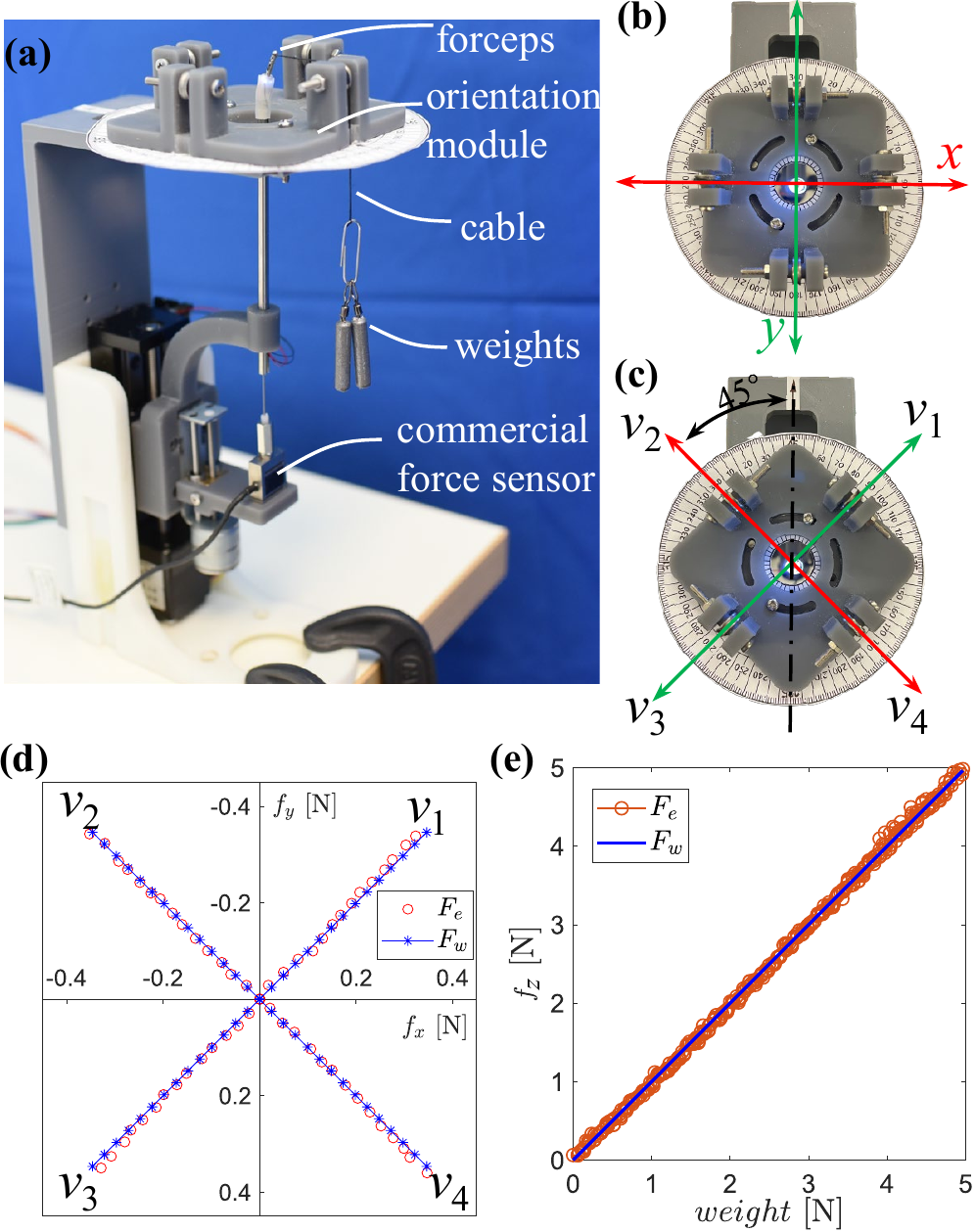}
		\caption{(a) The experimental setup for stiffness matrix calibration.
			The forceps were pulled by a cable connected with changeable weights.
			The orientation module was used to adjust the pull direction of the forceps.
			(b) shows the setup for calibration data collection.
			(c) shows the setup for verification data collection, where the orientation module was rotated 45$^\circ$ relative to that for calibration.
			$v_1$, $v_2$, $v_3$, and $v_4$ indicate the pulling directions for verification in $x$-$y$ panel.
			(d) and (e) show the verification results in $x$-$y$ panel and $z$-axial direction, respectively.
			$F_w$ and $F_e$ denote the force generated by the weight and the estimated result. 
		}
		\label{Figure:TFCalibration}	
	\end{figure}

	We used a series of weights to calibrate the haptics-enabled forceps' stiffness matrix, considering the influence of the driving cable and soft shell.
	The calibration platform is shown in Fig.~\ref{Figure:TFCalibration}(a).
	Here, the micro-level actuator is used to adjust the position and hold the forceps.
	An orientation module is installed concentrically to the forceps for adjusting the direction of the pulling force provided by a cable.
	This cable is connected to the forceps' jaws, and the force $F_w$ applied to it is adjusted by adding/removing weights.

	\subsubsection{Calibration and Verification}
	Two groups of data were collected for calibration and verification, respectively.
	When collecting the data for calibration in the $x$-$y$ panel, as shown in Fig.~\ref{Figure:TFCalibration}(b), the orientation module made the cable align with the $x$ and $y$ axis, and the force was increased from 0N to 0.49N at intervals of 0.07N.
	Then, for the $z$ axis, the driving force was increased from 0N to 5N at intervals of 0.5N.
	The target pose for each interval was recorded. 
	Then, substituting the calibration data to $\mathbf{K}_s = \mathbf{f}_s/\mathbf{d}_s$ in MATLAB, we obtained the stiffness matrix $\mathbf{K}_s$ as
	\begin{align}
		\mathbf{K}_s = \left[ 
		{\begin{array}{*{20}{c}}
				{0.9592}&{0.0932}&{-0.0184}\\
				{0.1210}&{0.8807}&{-0.0170}\\
				{0.0013}&{0.0012}&{3.8520}
		\end{array}} 
		\right] \, . \notag
		\label{equation:stiffness matrix}
	\end{align}
	
	To verify the calibrated $\mathbf{K}_s$, we rotated the orientation module for $45^\circ$, as shown in Fig.~\ref{Figure:TFCalibration}(c), and collected the data following the same procedure of calibration.
	For the $x$-$y$ panel the interval was set to 0.035N, while for the $z$ axis the interval was 0.02N. 
	The comparison of weights and estimated forces are plotted in Fig.~\ref{Figure:TFCalibration}(d) and (e), which show the results of the $x$-$y$ panel and the $z$-axial direction, respectively. 
	The absolute mean, maximum, and RMS errors of ($f_x$, $f_y$, $f_z$) in these three comparisons are ($0.0064$, $0.0045$, $0.0383$)N, ($0.0233$, $0.0154$, $0.1996$)N, and (0.8494, 0.5884, 4.9367)N, which are (1.86\%, 1.30\%, 0.76\%), (6.71\%, 4.46\%, 3.97\%), and (2.45\%, 1.70\%, 0.98\%) of the measured force amplitude (MFA), respectively.
	This indicates that the calibrated $\mathbf{K}_s$ can be used to estimate the force applied to the forceps.

\begin{figure}[t!]
	\centering
	\includegraphics[width=0.48\textwidth]{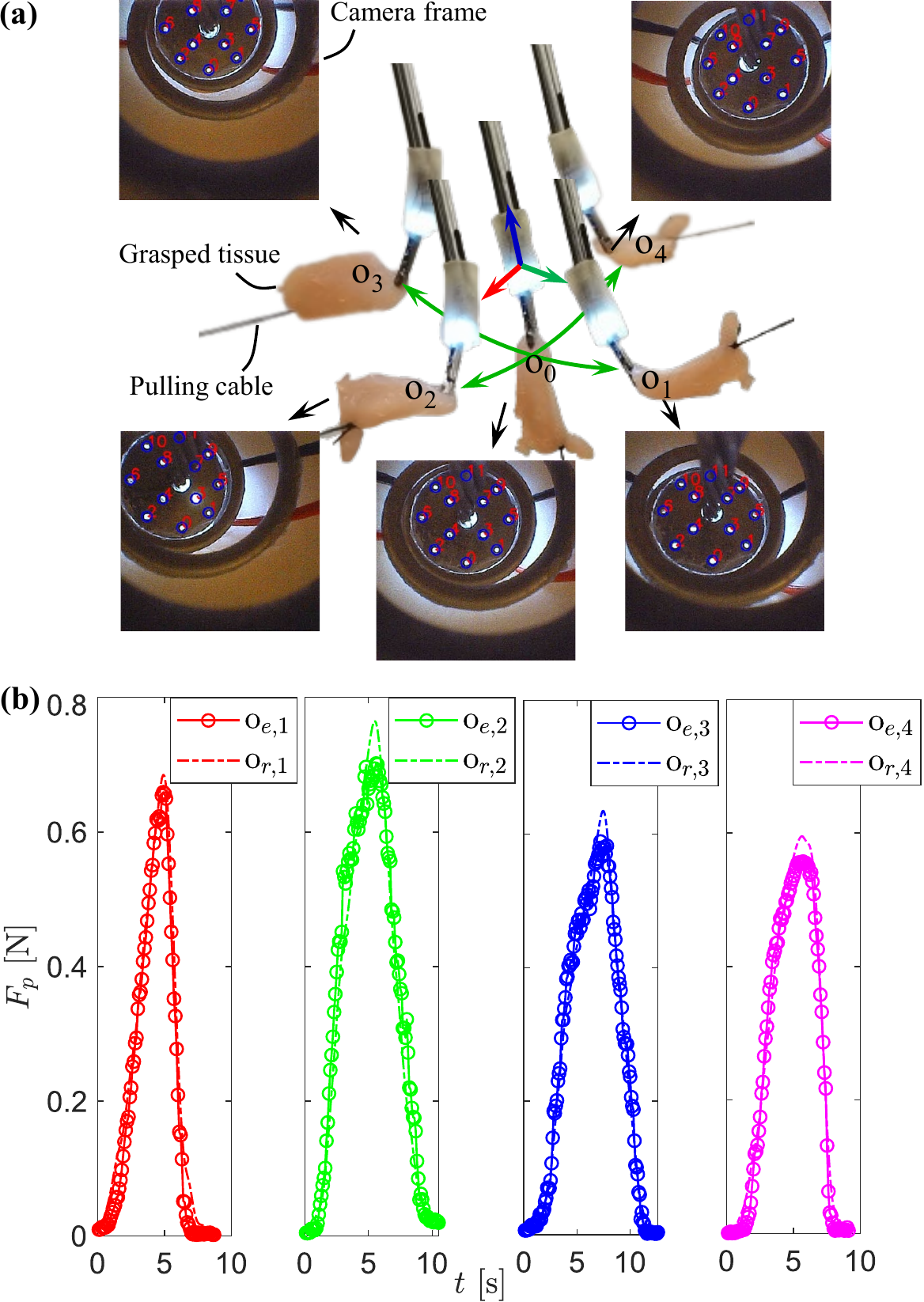}
	\caption{The evaluation of Mode-I ($F_p$ estimation) in various directions.
		(a) A grasped tissue was pulled by a cable from the initial state o$_0$ to $+y$ (o$_1$), $+x$ (o$_2$), $-y$ (o$_3$) and $-x$ (o$_4$).
		Snapshots show states when the spring has maximum deformation in different orientations, and the corresponding camera frames are attached below them.
		(b) $F_p$ estimated by our proposed sensing module (o$_{e,i}$) and that measured by the reference commercial sensor (o$_{r,i}$).}
	\label{Figure:Pulling_Orientations}	
\end{figure}
\begin{table}[t]
	\centering
	\caption{Errors of $F_p$ [N] estimation in various orientation}	
	\begin{tabular}{ccccc}
		\toprule
		Error &o$_{1}$ & o$_{2}$ & o$_{3}$ & o$_{4}$ \\ \toprule
\scriptsize{Mean}&{\scriptsize{0.030(4.41\%)}}&{\scriptsize{0.038(5.15\%)}}&{\scriptsize{0.028(4.01\%)}}&{\scriptsize{0.023(3.91\%)}}\\ \hline
\scriptsize{Max}& {0.077\scriptsize{(11.17\%)}} &{\scriptsize{0.131(17.75\%)}}&{\scriptsize{0.128(14.42\%)}}&{\scriptsize{0.064(10.69\%)}}\\ \hline
\scriptsize{RMS}&{\scriptsize{0.038(5.50\%)}}&{\scriptsize{0.049(6.69\%)}}&{\scriptsize{0.043(6.16\%)}}&{\scriptsize{0.028(4.74\%)}}\\
		\hline
	\end{tabular}
	\label{Table:Robotic_Pulling}
\end{table}	

\subsection{Evaluation of Mode-I: $F_p$ Estimation}
In this experiment, the grasped tissue was pulled in various directions by a cable connected to a force sensor.
Fig.~\ref{Figure:Pulling_Orientations}(a) shows that the grasped tissue is pulled from the initial state o$_0$ to $+y$ (o$_1$), $+x$ (o$_2$), $-y$ (o$_3$), and $-x$ (o$_4$).
Snapshots show the states when the flexure has maximum deformation in different orientations, and the corresponding camera registration frame are attached below them.
Fig.~\ref{Figure:Pulling_Orientations}(b) shows the forces estimated by the proposed sensing module (o$_{e,i}$) and that measured by the commercial sensor (o$_{r,i}$) connected to the tissue-pulling cable.
Table \ref{Table:Robotic_Pulling} lists the comparison result of $F_p$ between o$_{e,i}$ and o$_{r,i}$.
The absolute mean, maximum, and RMS errors of the four tests are under 0.038N (5.15\% of MFA), 0.131N (17.75\% of MFA), and 0.049N (6.69\% of MFA), which means the forceps are valid for estimating $F_p$ in various directions and meets the requirement of general surgical applications (for example, the clinical review \cite{golahmadi2021tool} indicated that an absolute error of less than 0.28N can meet the requirement for tissue retraction with grasping in general surgery).

\subsection{Evaluation of Mode-II: $F_g$ Estimation}
	\label{subsection:grasping_experiment}

	\subsubsection{Experimental Setup}
	Fig.~\ref{Figure:Grasping_setup}(a) shows the experimental setup, where a commercial pressure sensor (FlexiForce A201-1 lbs, Tekscan, USA), with a resolution of 0.02N, and a force sensor (ZNLBS-5kg, CHINO, China) were adopted as references.
	The pressure sensor was grasped by the forceps for direct measurement of the grasping force.
	Because the pressure sensor's sensing area was a 10mm-diameter circle, 3D-printed additional surfaces were attached to the forceps' jaws to ensure sufficient contact.
	The driving force reached around 7N during the grasping action in each experiment.
    Since, from (\ref{euquation:Fg}), we can see the grasping force $F_g$ is also related to pulling force $F_p$, we connected the pressure sensor to a commercial force sensor by a flexure to measure the applied pulling force.

	\subsubsection{Evaluations}
\begin{figure}[t]
	\centering
	\includegraphics[width=0.48\textwidth]{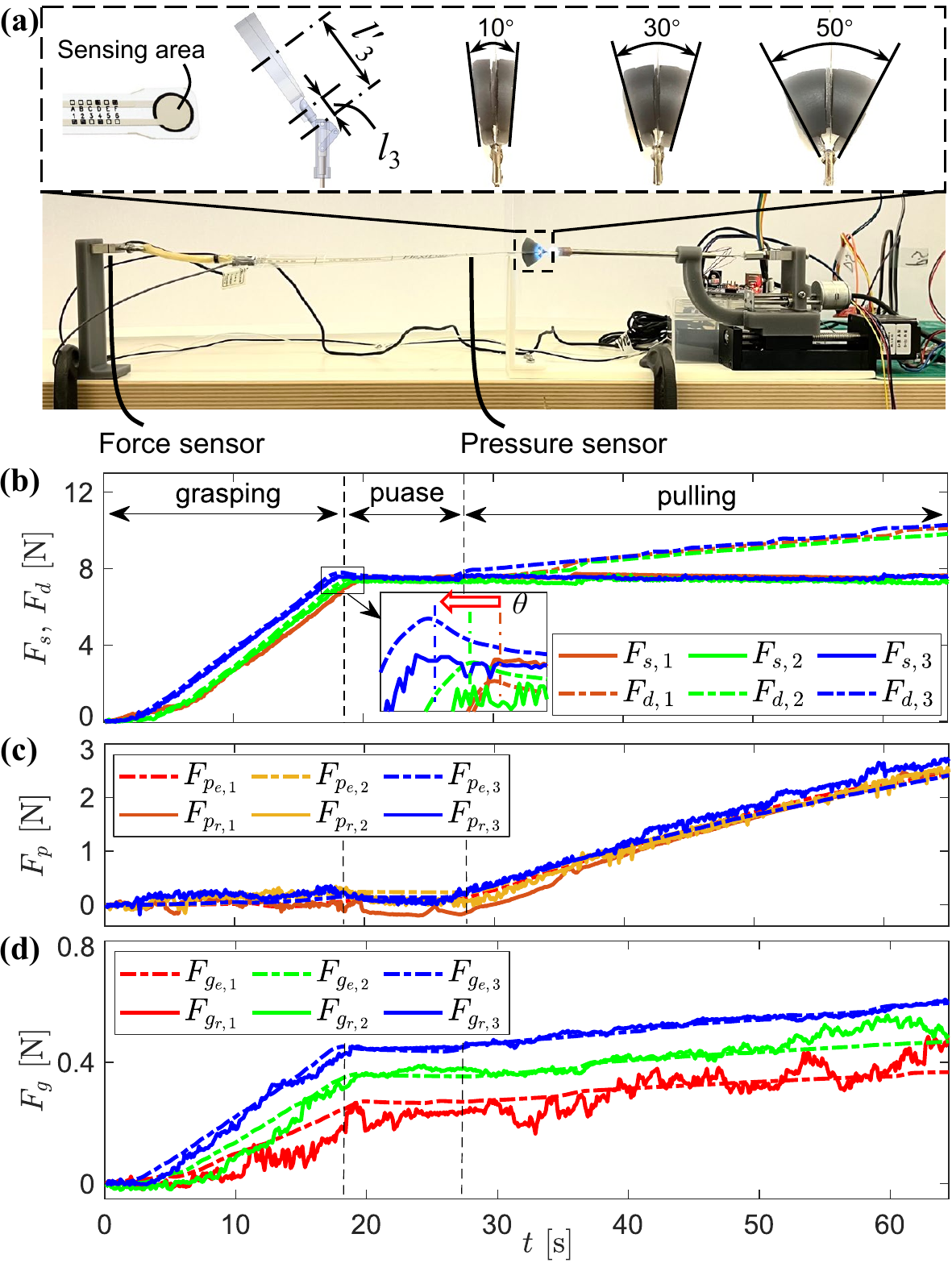}
	\caption{The evaluation of Mode-II ($F_g$ estimation).
		(a) The experimental setup, where a commercial pressure sensor and a force sensor were used as references for grasping and pulling forces, respectively.
		The inset shows three 3D-printed contact surfaces configured with $\theta$ = $10^\circ$, $30^\circ$ and $50^\circ$. 
		(b) shows $F_d$ provided by the driving cable and $F_s$ measured by our proposed force sensing module. Their difference is the pulling force $F_p$ applied to the grasped pressure sensor.
		Subscript $i \in$(1,2,3) denotes the $i$-th experiment that correspond to $\theta = 10^\circ$, $30^\circ$ and $50^\circ$.
		The inset shows that grasping time varies for different $\theta$, and the red arrow shows the increased direction of $\theta$. 
		(c) The comparison of pulling force $F_p$, where $F_{p_{e,i}}$ and $F_{p_{r,i}}$ denote the forces of the $i$-th experiment estimated by our proposed sensing module and that measured by the left commercial force sensor shown in (a), respectively.
		(d) The comparison of grasping force $F_g$, where $F_{g_{e,i}}$ and $F_{g_{r,i}}$ denote the forces of the $i$-th experiment estimated by our proposed sensing module and that measured by the reference commercial pressure sensor, respectively.
	}
	\label{Figure:Grasping_setup}	
\end{figure}

	We carried out three experiments with different $\theta$ configurations.
	Each experiment was presented as a continuous process, and the results are shown in Fig.~\ref{Figure:Grasping_setup}(b), (c), and (d), where the subscript $i\in$(1,2,3) denotes the $i$-th experiment that corresponds to  $\theta = 10^\circ$, $30^\circ$, and $50^\circ$.
	The forceps grasped the pressure sensor and reached the target grasping force during 0$\sim$18s.
	After a pause phase (18$\sim$26s), the forceps pulled the grasped sensor backward for 25mm in the pulling phase (26$\sim$65s). 

	Fig.~\ref{Figure:Grasping_setup}(b) depicts $F_d$ provided by the driving cable and $F_s$ measured by our proposed sensing module.
	$F_s$ and $F_d$ were almost equal in the grasping and pause phases.
	This is because the pulling force applied to the forceps was almost zero in these two phases, as there was no pulling action, and the pressure sensor was floating.
	On the contrary, in the pulling phase, $F_d$ increased while $F_s$ kept constant.
	This is because the pressure sensor has been pulled ($F_p = F_d - F_s$), but the flexure deformation kept constant. 
	The difference between $F_d$ and $F_s$ was the pulling force $F_p$ applied to the pressure sensor.
	As the three experiments followed the same procedure, their data almost overlay each other.
	However, because $\theta$ is configured differently, the time for performing the grasping is slightly different.
	The inset in Fig.~\ref{Figure:Grasping_setup}(b) shows that the bigger $\theta$ is, the quicker the grasping finishes.

	Fig.~\ref{Figure:Grasping_setup}(c) shows the commercial sensor measured pulling force $F_{p_{r,i}}$ in the experimental procedure, and we compared it with forceps estimated result $F_{p_{e,i}}$.
	We can see that the pulling force of these three experiments almost overlay each other as the pulling distances were equal.
	A comparison between the commercial sensor results $F_{p_r}$ and our estimations $F_{p_e}$ is listed in Table \ref{Table:Grasping_Pulling}. 
	The mean, maximum, and RMS errors of the three experiments are under 0.133N, 0.497N, and 0.167N, respectively.
	This also reflects that the forceps are valid for estimating the pulling force when they grasp tissue.

\begin{table}[t]
	\centering
	\caption{Errors of $F_g$ and $F_p$ estimation in grasping procedure}	
	\begin{tabular}{ccccc}
		\toprule
		Force & Error & $10^\circ$ & $30^\circ$ & $50^\circ$ \\ \toprule
		& Mean & {0.118\scriptsize{(4.81\%)}} & {0.094\scriptsize{(3.88\%)}} & {0.133\scriptsize{(5.50\%)}}\\ \cline{2-5}
		$F_p$ [N]& Max  & {0.299\scriptsize{(12.21\%)}} & {0.278\scriptsize{(11.48\%)}} & {0.497\scriptsize{(20.51\%)}}\\ \cline{2-5}
		& RMS   & {0.145\scriptsize{(5.94\%)}} & {0.112\scriptsize{(4.62\%)}} & {0.167\scriptsize{(6.87\%)}}\\ \hline
		
		& Mean &{0.040\scriptsize{(8.13\%)}} &{0.027\scriptsize{(4.90\%)}}&{0.012\scriptsize{(2.00\%)}}\\ \cline{2-5}
		$F_g$ [N]& Max &{0.133\scriptsize{(27.33\%)}} &{0.096\scriptsize{(17.37\%)}}&{0.049\scriptsize{(8.05\%)}}\\ \cline{2-5}
		& RMS  &{0.049\scriptsize{(10.03\%)}} &\scriptsize{0.036{(6.50\%)}}&{0.016\scriptsize{(2.67\%)}}\\
		\hline
	\end{tabular}
	\label{Table:Grasping_Pulling}
\end{table}
	
	Fig.~\ref{Figure:Grasping_setup}(d) compares the grasping force $F_g$, where $F_{g_{e,i}}$ and $F_{g_{r,i}}$ denote the forces of $i$-th experiment estimated by our proposed forceps and commercial pressure sensor, respectively.
	Because of the additional surfaces that are shown in the inlet of Fig.~\ref{Figure:Grasping_setup}(a), $l_3$ in (\ref{euquation:Fg}) was set as $l^{\prime}_3$ in these experiments.
	According to Fig.~\ref{Figure:Grasping_setup}(d), we can see that when the driving force $F_d$ is equal, the grasping force $F_g$ is positively related to $\theta$.
	This phenomenon conforms with the force calculation method (\ref{equation:fp}) and (\ref{euquation:Fg}), as the increased $F_p$ could entail greater $F_d$ and then result in increased $F_g$.
	Table \ref{Table:Grasping_Pulling} lists the absolute mean, maximum, and RMS errors, and their percentage of MFA, between $F_{g_r}$ and $F_{g_e}$ of the three experiments. 
	Their values are under 0.040N, 0.133N, and 0.049N, respectively.
	This indicates that the forceps are valid for estimating the grasping force and meet the requirements of surgical applications \cite{golahmadi2021tool}.

\subsection{Ex vivo Robotic Experiments}
\label{Subsection:Dynamic_Experiment}
\subsubsection{Experimental Setup}
\begin{figure}[t!]
		\centering
		\includegraphics[width=0.45\textwidth]{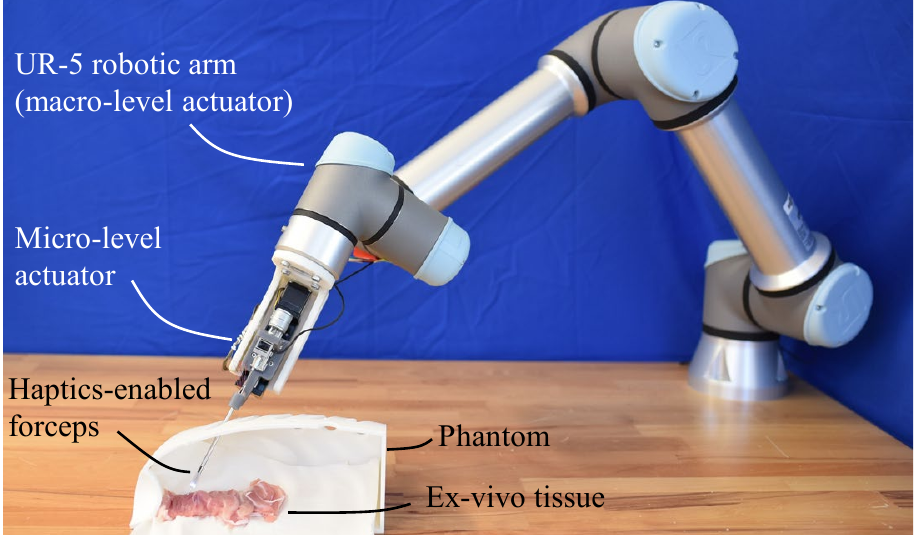}
		\caption{The setup of robotic experiments. 
			A 3D-printed human body phantom was used for placing \textit{ex vivo} chicken tissue that simulated the lesion.
			The instrument was implemented through a simulated MIS port on the phantom.}
		\label{Figure:Robotic_setup}	
	\end{figure}

To further verify the feasibility of the proposed system, we conducted a group of robotic experiments on an \textit{ex vivo} tissue.
Fig.~\ref{Figure:Robotic_setup} shows the experimental setup, where a \textit{UR}-5 robotic arm was adopted as the macro-level actuator. 
The \textit{ex vivo} chicken tissue was placed in a human body phantom to simulate the lesion, and the instrument was implemented through a simulated minimally invasive port on the phantom.

	\subsubsection{Automatic Tissue Grasping}
\begin{figure*}[t!]
		\centering
		\includegraphics[width=0.98\textwidth]{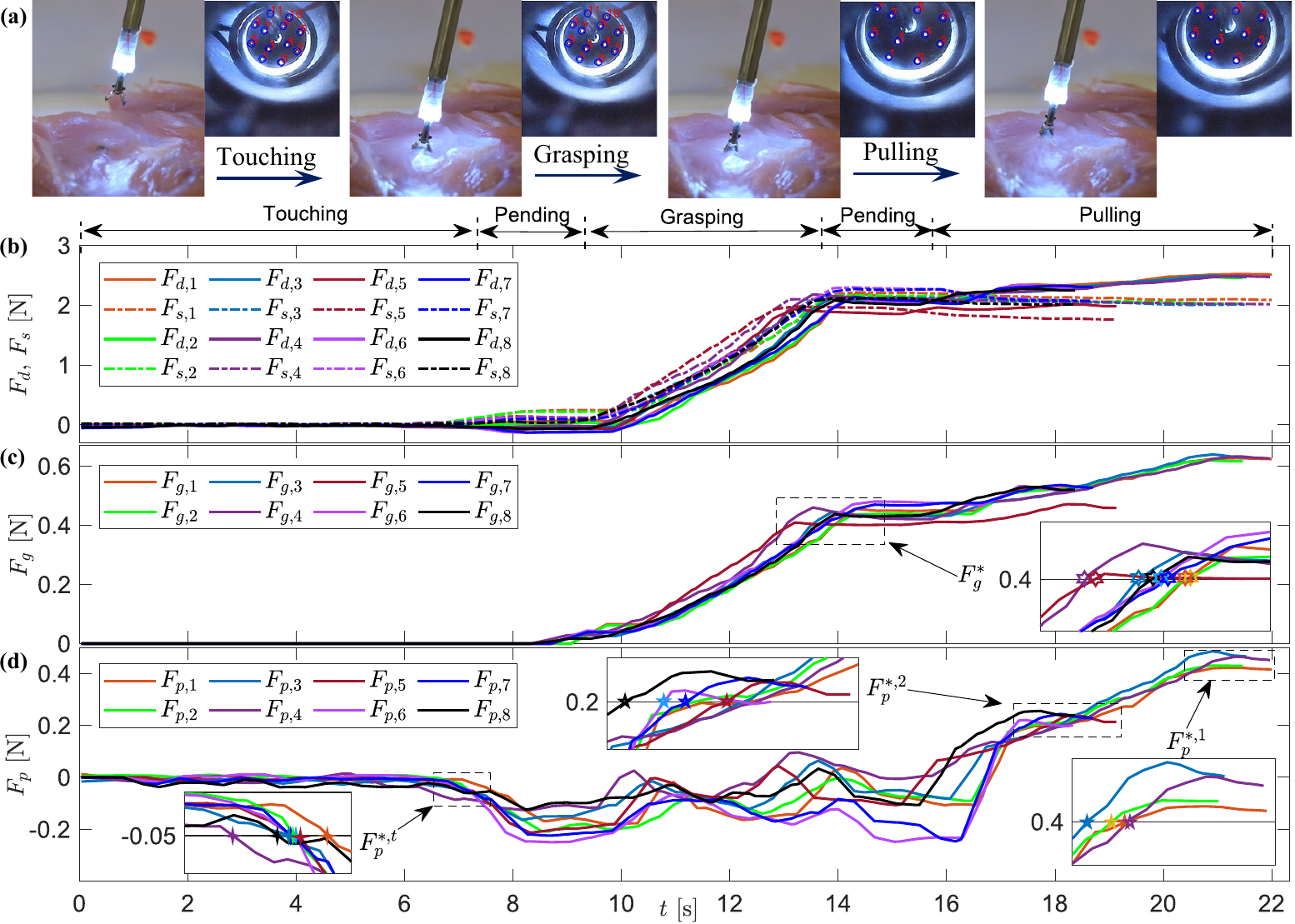}
		\caption{The results of automatic tissue grasping experiments, where experiments can be classified into two groups with two different target grasping forces $F_p^{\ast}$. 
			Experiments of the first group are marked with footnotes 1, 2, 3, and 4, while those of the second are marked with footnotes 5, 6, 7, and 8.
			(a) Key snapshots of the experimental procedure are attached with target tracking results. 
			(b) shows the measured driving force $F_d$ and the estimated elastic force $F_s$ of these experiments.
			Plot (b) also shows the experimental phases.
			Touching means the forceps touch the targeted tissue in the opened form.
			Pending means the speeds of the system's motors have been set to zero for waiting.
			Grasping means the forceps grasp the touched tissue. 
			Pulling means the forceps pull the grasped tissue.
			(c) shows the estimated $F_g$, where $F_g^{\ast}$=0.4N is the targeted grasping force for grasping action.
			The partial enlargement shows each experiment has reached $F_g^{\ast}$ successfully and is marked with a hexagon.   
			(d) shows the estimated $F_p$, where $F_t^{\ast}$=-0.05N is the threshold for touch detection, while $F_p^{\ast,1}$=0.4N and $F_p^{\ast,2}$=0.2N are two target pulling forces for the first and second group experiments, respectively.
			The left partial enlargement shows the touching detection and is marked with a four-pointed star for each experiment.
			The two right partial enlargements show the reaching of $F_p^{\ast}$ of two group experiments and are marked with the pentagram.
		}
		\label{Figure:Robotic_grasping}	
	\end{figure*}

	We carried out automatic tissue grasping procedure with one targeted grasping force $F_g^{\ast}$ and two different targeted pulling forces $F_p^{\ast,1}$ and $F_p^{\ast,2}$.
	The key experimental scenes and results are shown in Fig.~\ref{Figure:Robotic_grasping}, and the experiments can be roughly divided into five phases.
	Transformations between these phases were automatically performed depending on the grasping force $F_g$ and pushing/pulling force $F_p$ estimated by the haptics-enabled forceps.
	
	With Mode-I enabled, the forceps were driven to touch the tissue during the touching phase (0$\sim$7.5s) until $F_p$ reached the touching detection threshold $F_p^{\ast,t}$ followed by a short pending period (7.5$\sim$9.5s). 
	Then, Mode-II was enabled, and the forceps performed the tissue grasping in the grasping phase (9.5$\sim$13.5s) until $F_g$ reached the targeted value $F_g^{\ast}$. 
	Finally, with another short pending period (13.5$\sim$15.5s), the grasped tissue was pulled up and held for a while with targeted pulling force $F_p^{\ast}$.
	Because the targeted pulling force $F_p^{\ast}$ for the first and second group experiments were different, the time consumed for pulling varied. 
	For the first group with $F_p^{\ast,1}=0.4$N, the period was around 6s, while for the second group with $F_p^{\ast,2}=0.2$N, it was around 2s.
	
	Fig.~\ref{Figure:Robotic_grasping}(b) shows the measured driving force $F_d$ and the estimated elastic force $F_s$ of these experiments.
	Fig.~\ref{Figure:Robotic_grasping}(c) shows the estimated $F_g$, where $F_g^{\ast}=0.4$N is the targeted grasping force for grasping action. 
	Fig.~\ref{Figure:Robotic_grasping}(d) shows the estimated $F_p$, where $F_g^{\ast,t}=-0.05$N is the threshold for touch detection, while $F_p^{\ast,1}=0.4$N and $F_p^{\ast,2}=0.2$N are two target pulling forces for the first and second group experiments, respectively.  
	According to the experiment results, we can see that the haptics-enabled forceps can be implemented in robotic surgery for multi-modal force sensing.
	We also noticed that, during the grasping process, the pulling force $F_p$ varies notably, which can potentially indicate successful grasping.
	
	\subsection{Discussion}
	The accuracy, sampling frequency, and resolution of the sensing module can be reconfigured with different cameras and flexures, which can be defined according to the surgical task requirements.
	Currently, referring to \cite{keller2011equivalent}, we linearly modeled the integrated haptics-enabled forceps with the calibrated $\mathbf{K}_s$ to have a simple calculation complexity with acceptable accuracy. 
	The accuracy can potentially be improved with a non-linear model, which can be obtained by more advanced calibrations and fitting methods.  
	The sampling frequency of the proposed sensing module is equal to the camera rate, 30Hz in the current prototype, and no filter has been applied to the collected data yet. 
	The sensing resolution depends on the camera's resolution and the spring's stiffness, and that of the prototype is 0.001N in the $x$-$y$ panel and 0.005N in the $z$-axis.
	The resolution in the $x$-$y$ panel is higher than in the $z$-axis because the camera is more sensitive in the $x$-$y$ panel displacement \cite{ouyang2020low}, and the spring stiffness is higher in the $z$-axis.
	To reduce the anisotropy in resolution, a hemisphere-shaped target is potentially helpful \cite{fernandez2021visiflex}.
 	Moreover, to ensure the feasibility of closed-loop force control, the filter to improve accuracy and smoothness should also be studied carefully.
	  
	Since the adopted sensing method relies on the spring's deformation, estimated by cameras tracking the target mounted on the opposite end of the spring, the sensing module is limited to estimating forces where the applied location ranges from the target to the forceps' tip.
	Additionally, this paper focuses on estimating the multi-modal forces when the forceps pull/push and grasp tissues, and only the displacement $\mathbf{d}_s$ is utilized for force estimation.
	Because the estimated transformation ${}^{M_0}_{M_t}\mathbf{T}$ also contains the target's current pose information, the wrench is potentially estimable \cite{ouyang2020low,fernandez2021visiflex} and will be considered in future works.

	Additionally, the proposed haptics-enabled forceps have minimized tool-side circuitry including only two micro-sized LEDs and an endoscopy, while the power supply and signal processing circuits are integrated at the proximal end and are removable, and the adopted camera can be sterilized repeatably, such as Ethylene oxide and STERRAD sterilization \cite{rutala2008guideline}. This improves the haptics-enabled forceps' potential to be sterilized repeatedly as well.
    
	Although in this paper, the biopsy forceps are adopted as an example, the presented sensing module can be integrated into a wide range of forceps with similar structures. 
	The sensing module can probably be used as an independent sensor for other surgical applications, for example, palpation.
	Moreover, inspired by the robotic experiments, the successful grasp can potentially be indicated by the variation of $F_p$ during each grasping process, which will be investigated in future works.

	
	\section{Conclusion}
	\label{Section:conclusion}
	This paper presented a vision-based force sensing module that is adaptive to micro-sized biopsy forceps.
	An algorithm was designed to calculate the force applied to the sensing module with a registration method for tracking and estimating the pose of the sensing module's target.
	Integrating the developed sensing module into the biopsy forceps, in conjunction with a single-axis force sensor at the proximal end, a haptics-enabled forceps was further proposed.
	Mathematical equations were derived to estimate the multi-modal force sensing of the haptics-enabled forceps, including pushing/pulling force (Mode-I) and grasping force (Mode-II).
	The methods for estimating multi-modal forces were presented with experimental verification.
	Groups of automatic robotic \textit{ex vivo} tissue grasping procedures were conducted to further verify the feasibility of the proposed sensing method and forceps. 
	The results show that the proposed method can enable multi-modal force sensing of the micro-sized forceps, and the haptics-enabled forceps are potentially beneficial to automatic tissue manipulation in operations such as thyroidectomy, ENT surgery, and laparoscopic surgery.
	The forceps will be integrated into a dual-arm robotic surgical system to further study the benefits of multi-modal force sensing for MIS.

	\bibliographystyle{IEEEtran}
	\bibliography{reference_haptic}

	
	\begin{IEEEbiography}[
		{\includegraphics[width=1in,height=1.25in,clip,keepaspectratio]{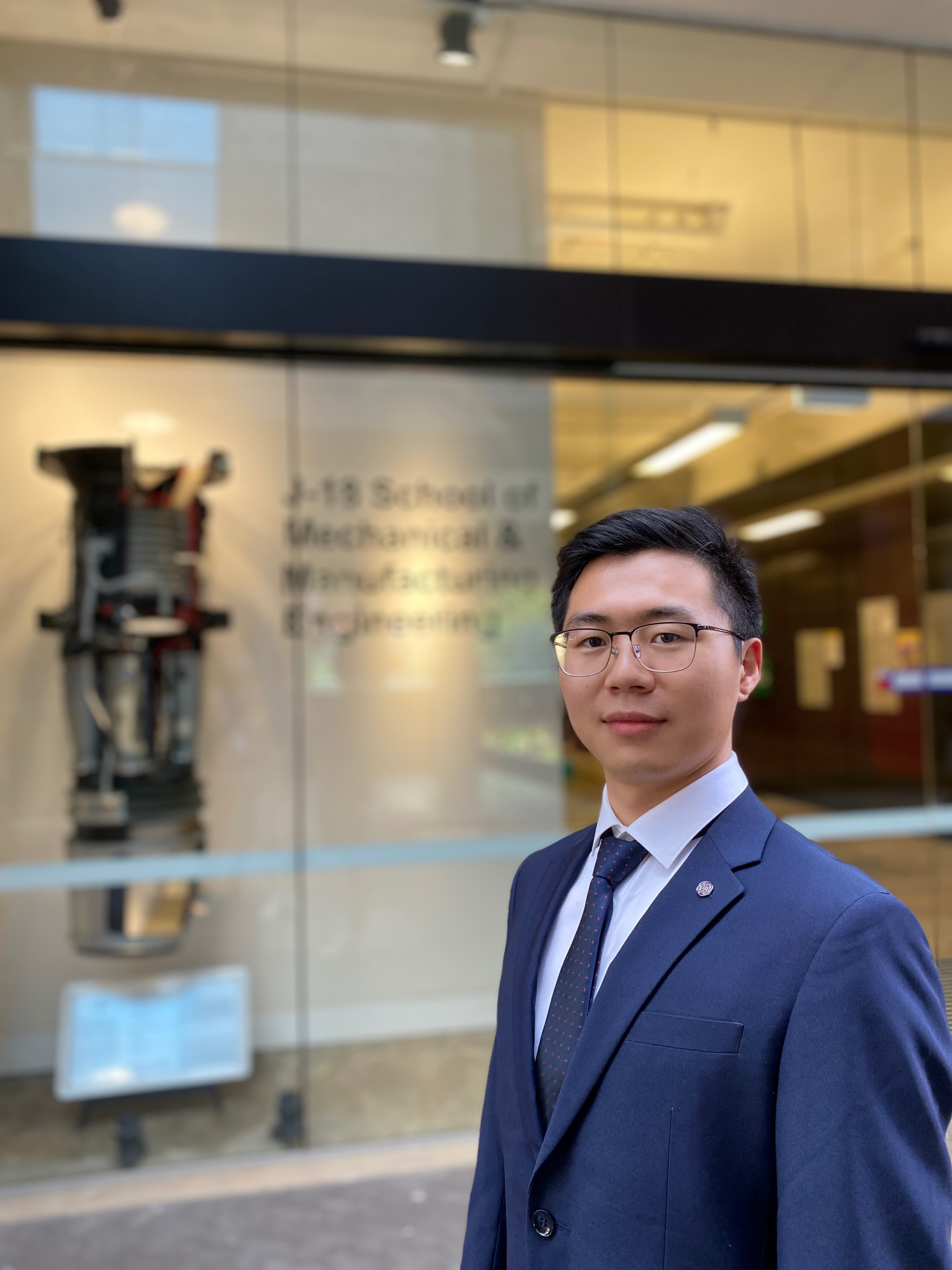}}
		]{Tangyou Liu} received the B.S. degree in mechanical engineering from the Southwest University of Science and Technology (SWUST) in 2017, and the M.S. degree in mechanical engineering as an outstanding graduate from the Harbin Institute of Technology, Shenzhen (HITsz) in 2021, supervised by Prof. Max Q.-H.Meng. Then, he worked in KUKA, China, as a system development engineer and was awarded the champion of the KUKA China R\&D Innovation Challenge. He was awarded the IEEE ICRA2023 Best Poster. He is pursuing his Ph.D. degree in mechatronic engineering at the University of New South Wales (UNSW), Sydney, Australia, under the supervision of Dr. Liao Wu. His current research interests include medical and surgical robots.
	\end{IEEEbiography}
	
	\begin{IEEEbiography}[
		{\includegraphics[width=1in,height=1.25in,clip,keepaspectratio]{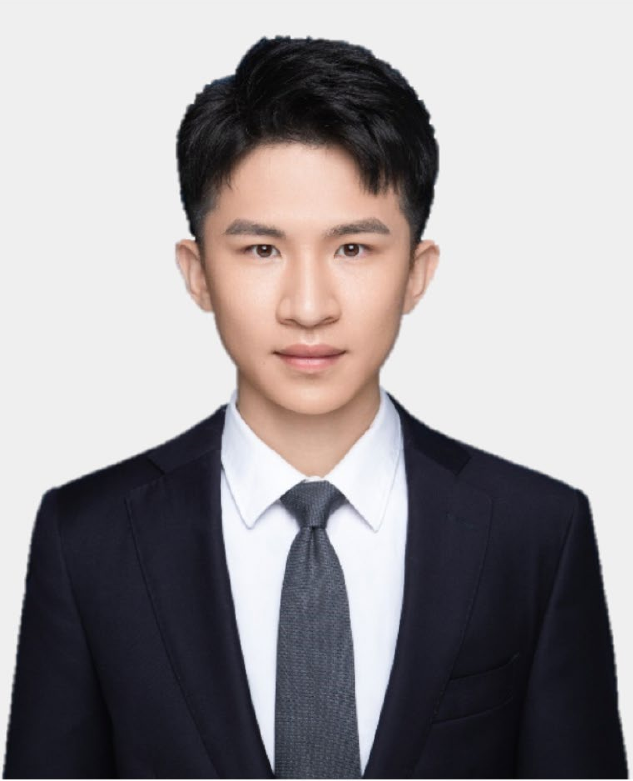}}
		]{Tinghua Zhang} received the B.S. and M.S. degrees in Mechanical Engineering from Jimei University (JMU) in 2020 and Harbin Institute of Technology (Shenzhen) (HITSZ) in 2023, respectively. Now, he is working at The Chinese University of Hong Kong (CUHK) as a research assistant.
		His research interest is mainly in medical robotics, including robotic OCT, optical tracking and force sensing. He was awarded the best student paper finiallist of IEEE ROBIO 2022.
	\end{IEEEbiography}

	\begin{IEEEbiography}[
	{\includegraphics[width=1in,height=1.25in,clip,keepaspectratio]{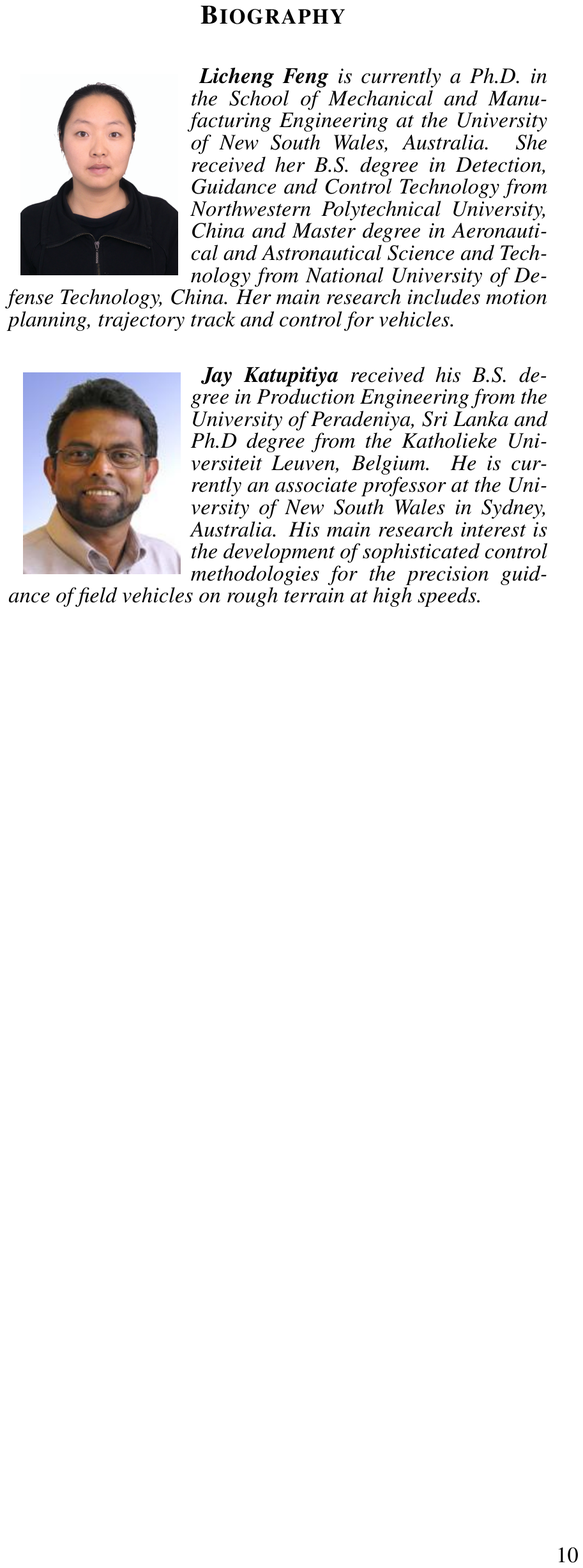}}
	]{Jay Katupitiya} received his B.S. degree in Production Engineering from the University of Peradeniya, Sri Lanka and Ph.D degree from the Katholieke Universiteit Leuven, Belgium. He is currently an associate professor at the University of New South Wales in Sydney, Australia. His main research interest is the development of sophisticated control methodologies for the precision guidance of field vehicles on rough terrain at high speeds.
\end{IEEEbiography}
	
	\begin{IEEEbiography}[
		{\includegraphics[width=1in,height=1.25in,clip,keepaspectratio]{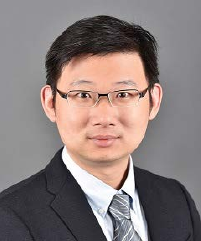}}
		]{Jiaole Wang} received the B.E. degree in mechanical engineering from Beijing Information Science and Technology University, Beijing, China, in 2007, the M.E. degree from the Department of Human and Artificial Intelligent Systems, University of Fukui, Fukui, Japan, in 2010, and the Ph.D. degree from the Department of Electronic Engineering, The Chinese University of Hong Kong (CUHK), Hong Kong, in 2016. He was a Research Fellow with the Pediatric Cardiac Bioengineering Laboratory, Department of Cardiovascular Surgery, Boston Children’s Hospital and Harvard Medical School, Boston, MA, USA. He is currently an Associate Professor with the School of Mechanical Engineering and Automation, Harbin Institute of Technology, Shenzhen, China. His main research interests include medical and surgical robotics, image-guided surgery, human-robot interaction, and magnetic tracking and actuation for biomedical applications.
	\end{IEEEbiography}

	\begin{IEEEbiography}[
	{\includegraphics[width=1in,height=1.25in,clip,keepaspectratio]{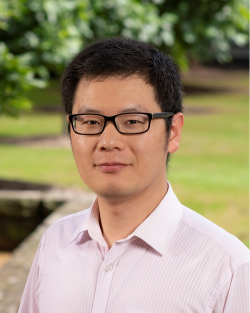}}
	]{ Liao Wu} is a Senior Lecturer with the School of Mechanical and Manufacturing Engineering, University of New South Wales, Sydney, Australia. He received his B.S. and Ph.D. degrees in mechanical engineering from Tsinghua University, Beijing, China, in 2008 and 2013, respectively. From 2014 to 2015, he was a Research Fellow at the National University of Singapore. He then worked as a Vice-Chancellor’s Research Fellow at the Queensland University of Technology, Brisbane, Australia from 2016 to 2018. Between 2016 and 2020, he was affiliated with the Australian Centre for Robotic Vision, an ARC Centre of Excellence. He has worked on applying Lie groups theory to robotics, kinematic modeling and calibration, etc. His current research focuses on medical robotics, including flexible robots and intelligent perception for minimally invasive surgery.
	\end{IEEEbiography}
	
\end{document}